\pdfoutput=1

\documentclass[11pt]{article}
\usepackage{amssymb} 
\usepackage{multirow}
\usepackage{graphicx}
\usepackage{array}
\usepackage{enumitem}
\usepackage{booktabs}  
\usepackage{makecell}  
\usepackage{pifont}
\usepackage[table]{xcolor}
\usepackage{colortbl}
\usepackage{hyperref}
\usepackage{amsmath}
\usepackage{arydshln}

\usepackage{tabularx}
\usepackage{xcolor} 
\usepackage[most]{tcolorbox}
\usepackage{lmodern}
\usepackage{setspace}
\usepackage{longtable} 

\usepackage{color}
\usepackage{jumoline}
\usepackage{proofread}

\usepackage[final]{acl}

\usepackage{times}
\usepackage{latexsym}

\usepackage[T1]{fontenc}

\usepackage[utf8]{inputenc}

\usepackage{microtype}

\usepackage{inconsolata}

\usepackage{graphicx}

\title{KokoroChat: A Japanese Psychological Counseling Dialogue Dataset Collected via Role-Playing by Trained Counselors}

\author{
 \textbf{Zhiyang Qi\textsuperscript{1}},
 \textbf{Takumasa Kaneko\textsuperscript{1}},
 \textbf{Keiko Takamizo\textsuperscript{2,3,4}},\\
 \textbf{Mariko Ukiyo\textsuperscript{2,3,4}},
 \textbf{Michimasa Inaba\textsuperscript{1,2}}
\\
 \textsuperscript{1}The University of Electro-Communications,\\
 \textsuperscript{2}Rapport Technologies, Inc.\\
 \textsuperscript{3}iDEAR Human Support Service,\\
 \textsuperscript{4}Japanese Organization of Mental Health and Educational Agencies
\\
 \texttt{\{qizhiyang,m-inaba\}@uec.ac.jp}
}

\begin{document}
\maketitle
\begin{abstract}

Generating psychological counseling responses with language models relies heavily on high-quality datasets. Crowdsourced data collection methods require strict worker training, and data from real-world counseling environments may raise privacy and ethical concerns. While recent studies have explored using large language models (LLMs) to augment psychological counseling dialogue datasets, the resulting data often suffers from limited diversity and authenticity. To address these limitations, this study adopts a role-playing approach where trained counselors simulate counselor-client interactions, ensuring high-quality dialogues while mitigating privacy risks. Using this method, we construct \textbf{KokoroChat}, a Japanese psychological counseling dialogue dataset comprising 6,589 long-form dialogues, each accompanied by comprehensive client feedback. Experimental results demonstrate that fine-tuning open-source LLMs with KokoroChat improves both the quality of generated counseling responses and the automatic evaluation of counseling dialogues. The KokoroChat dataset is available at \url{https://github.com/UEC-InabaLab/KokoroChat}.

\end{abstract}

\section{Introduction}

\begin{figure}[t!]
  \centering
  \includegraphics[width=\linewidth]{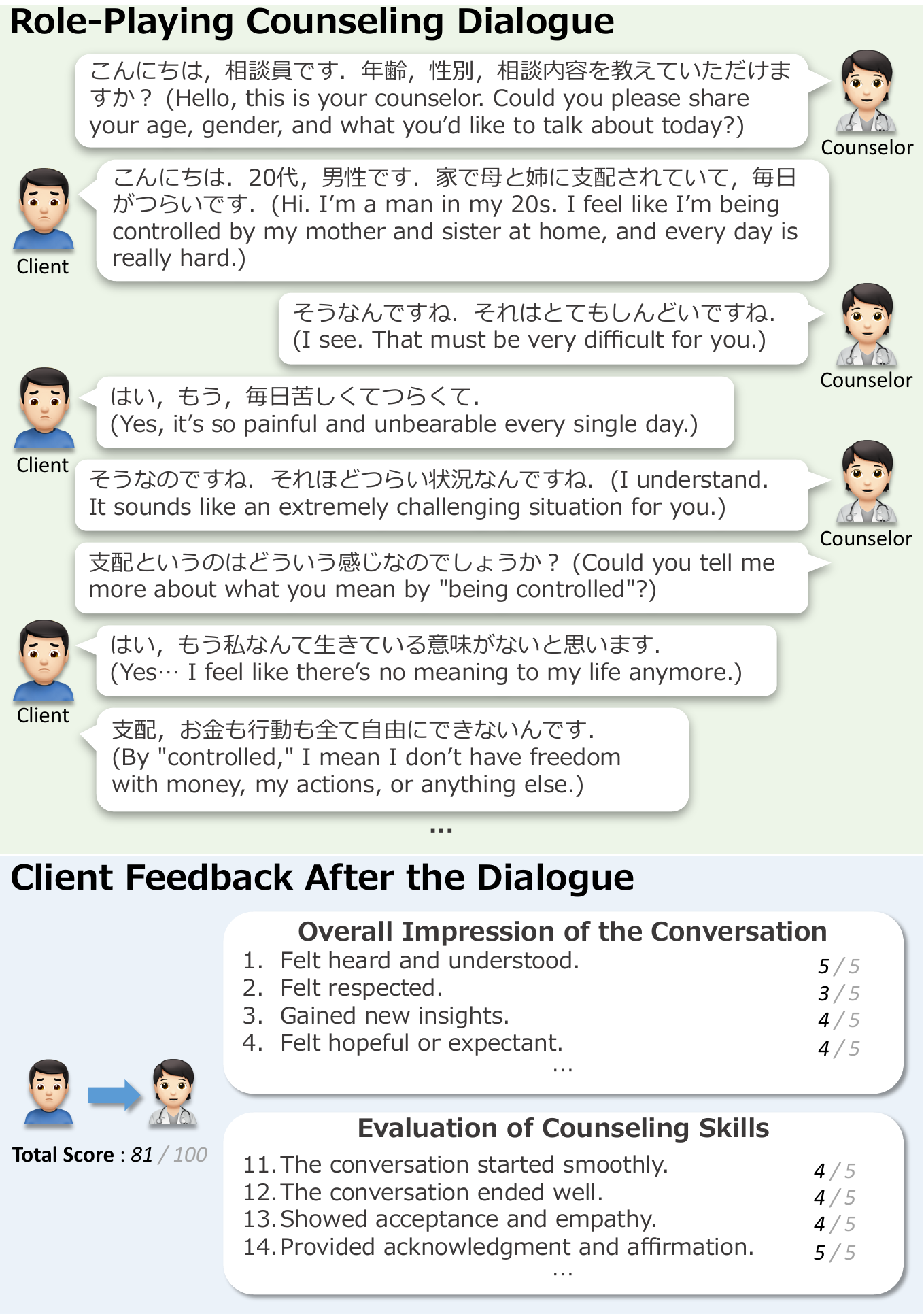}
  \caption{Each \textbf{KokoroChat} sample includes a counseling dialogue and client feedback, with both roles played by trained counselors.}
  \label{fig:fig_intro}
\end{figure}

Psychological issues have long posed a significant global challenge, with many individuals suffering from mental health disorders \cite{world2022mental}. However, limited medical resources restrict access to professional psychological counseling for most people \cite{samhsa2015nsduh}. To address this gap, researchers have explored language models for generating empathetic responses to provide emotional support. Advancing this research depends on constructing high-quality datasets. For instance, \citet{liu-etal-2021-towards} developed the ESConv dataset by training crowdworkers in emotional support skills, while \citet{li-etal-2023-understanding} created the Client-Reactions dataset by establishing an online mental health support platform to collect dialogues between real clients and professional counselors.
Despite these efforts, several challenges remain. Psychological counseling is a highly specialized form of communication \cite{skillful}, making it costly and time-consuming to train crowdworkers without professional backgrounds. Meanwhile, dialogue participants may struggle to fully grasp the experiences of individuals with mental disorders, making it difficult to simulate authentic interactions. Additionally, real counseling data may involve privacy and ethical concerns. As a result, the manual collection of such data often faces practical limitations.

\begin{table*}[t!]
\centering
\small
\renewcommand{\arraystretch}{1.2} 
\setlength{\tabcolsep}{5pt} 
\begin{tabular}{lccccccc} 
\toprule
\textbf{Dataset} & \textbf{Human-made} & \textbf{Score} & \textbf{Score items} & \textbf{Language} & \textbf{\# Dialogues} & \textbf{Avg. utterances} \\
\midrule
HealMe \cite{xiao-etal-2024-healme}        & \ding{55} & \ding{55} & -  & English & 1,300  & 6.0  \\ 
ESD-CoT \cite{zhang-etal-2024-escot}       & \ding{55} & \ding{55} & -  & English & 1,708  & 23.4  \\ 
C\textsc{actus} \cite{lee-etal-2024-cactus}  & \ding{55} & \ding{55} & -  & English & 31,577 & 31.5  \\  
SMILECHAT \cite{qiu-etal-2024-smile}       & \ding{55} & \ding{55} & -  & Chinese & 55,165 & 33.2  \\  
A\textsc{ug}ESC \cite{zheng-etal-2023-augesc}  & \ding{55} & \ding{55} & -  & English & 65,000 & 26.7  \\ 
\noalign{\vskip 3pt} 
\cdashline{1-7} 
\noalign{\vskip 3pt} 
Anno-MI \cite{AnnoMi}                      & \ding{51} & \ding{55} & -  & English & 133    & 72.9  \\
ESConv \cite{liu-etal-2021-towards}        & \ding{51} & \ding{51} & 2  & English & 1,300  & 29.5  \\  
Client-Reactions \cite{li-etal-2023-understanding} & \ding{51} & \ding{51} & 4  & Chinese & 2,382  & 78.5  \\  
\textbf{KokoroChat}                        & \ding{51} & \ding{51} & \textbf{20} & \textbf{Japanese} & \textbf{6,589} & \textbf{91.2}  \\  
\bottomrule
\end{tabular}
\caption{\label{tab:dataset_comparison}Comparison of psychological counseling datasets: LLM-augmented (top), human-collected (bottom).}
\end{table*}


Recently, LLMs have made significant strides in natural language generation and have shown considerable potential in generating psychological counseling responses \cite{inaba2024}. Consequently, many studies have leveraged LLMs for self-chat, rewriting, or mimicking existing datasets to construct or expand psychological counseling dialogue datasets \cite{zheng-etal-2023-augesc, xiao-etal-2024-healme, zhang-etal-2024-escot, lee-etal-2024-cactus, qiu-etal-2024-smile}. However, despite LLMs’ strong ability to generate psychological counseling responses, these augmented datasets often exhibit redundancy and homogeneity, leading to a lack of dialogue diversity \cite{zheng-etal-2024-self}. Moreover, as shown in Table~\ref{tab:dataset_comparison}, augmented datasets contain significantly fewer utterances than human-collected psychological counseling dialogues (e.g., Client-Reactions: 78.5), limiting their effectiveness and applicability.


To address these issues, this study employs a role-playing approach for data collection, in which trained professional and trainee counselors simulate interactions between a counselor and a client. Compared to traditional crowdsourced methods, this approach ensures higher professionalism and dialogue quality, as both participants are trained counselors. Additionally, unlike direct collection of real counseling dialogues, this method mitigates privacy and ethical risks. Furthermore, compared to LLM-augmented datasets, the dialogues collected through this approach offer stronger assurances of professionalism and authenticity.


Through a role-playing approach, we develop \textbf{KokoroChat}, a high-quality dialogue dataset for psychological counseling.
As shown in Figure~\ref{fig:fig_intro}, trained professional and trainee counselors play both counselor and client roles, engaging in approximately one-hour online text-based counseling sessions.
Considering that an objective and detailed scoring mechanism can quantify counseling quality and support counselor skill development, we collect client feedback after each session.
The client-role player evaluates the counselor-role player based on two key dimensions: overall impressions and professional skills. The evaluation consists of 20 assessment items, each rated on a scale of 0 to 5, with a maximum total score of 100.
This study involves 480 participants, all of whom have completed 10 hours of training in online text-based psychological counseling. Over one-third are professional counselors, while the remaining participants are trainees who have studied relevant topics for six months to one year and aspire to become certified counselors.

As shown in Table~\ref{tab:dataset_comparison}, we collect 6,589 high-quality psychological counseling dialogues, averaging 91.2 utterances per dialogue, each accompanied by detailed client feedback. To our knowledge, KokoroChat is the largest human-collected psychological counseling dialogue dataset to date, with session durations aligning with real-world counseling practices (approximately one hour per session). 
Notably, KokoroChat is a Japanese-language dataset, offering linguistic resources for psychological counseling dialogue research from a diverse cultural perspective. Given the global demand for psychological counseling, developing datasets that encompass multiple cultures and languages is essential for enhancing models’ adaptability to users from different backgrounds. By filling the gap in Japanese psychological counseling dialogue data, KokoroChat provides a foundation for cross-cultural research in psychological counseling.


To summarize: (1) we develop KokoroChat, the largest manually collected psychological counseling dialogue dataset to date, using a role-playing approach, with detailed client feedback;
(2) we fine-tune an open-source LLM to demonstrate that KokoroChat enhances LLM performance in generating psychological counseling responses;
(3) we train a dialogue evaluation model using client feedback from KokoroChat, and our experimental results show that this model provides more robust and accurate evaluation outcomes.



\section{Related Work}

\subsection{Psychological Counseling in NLP}

In recent years, psychological counseling has attracted significant attention in the field of natural language processing (NLP). Some studies have focused on generating empathetic responses, where systems provide appropriate feedback by understanding users’ emotions \cite{rashkin-etal-2019-towards, sharma-etal-2020-computational, zheng-etal-2021-comae}. However, empathy alone is insufficient to address the complex demands of psychological counseling. To bridge this gap, \citet{liu-etal-2021-towards} proposed the Emotional Support Conversation (ESC) task, which requires systems not only to exhibit empathy but also to deeply explore users' concerns and offer effective guidance to help them navigate challenges.

As LLMs' generation capabilities advance, their potential applications in psychological counseling have gained further attention. For example, \citet{inaba2024} demonstrated that GPT-4’s \cite{gpt4} responses in psychological counseling scenarios are comparable to those of professional counselors. Additionally, several LLM-based psychological counseling chatbots, such as ChatCounselor \cite{ChatCounselor}, MeChat \cite{qiu-etal-2024-smile}, and SoulChat \cite{chen-etal-2023-soulchat}, have emerged. These systems are typically fine-tuned on manually curated or LLM-augmented psychological counseling data to adapt to specific scenarios.
However, the field still faces challenges, particularly the lack of high-quality, diverse professional datasets. To address this, our study develop KokoroChat, a high-quality psychological counseling dialogue dataset, through role-playing by professional counselors. This dataset aims to facilitate the development of psychological counseling dialogue systems.

\subsection{Counseling Dialogue Datasets}

The key to equipping language models with psychological counseling capabilities lies in the availability of high-quality datasets. Currently, psychological counseling datasets fall into two main categories: manually constructed datasets and those generated by LLMs.
Manually constructed datasets typically consist of human dialogues. For example, Anno-MI \cite{AnnoMi} was developed by extracting full motivational interview dialogues from online videos. ESConv \cite{liu-etal-2021-towards} collected emotional support dialogues from crowdworkers trained in specialized skills, while Client-Reactions \cite{li-etal-2023-understanding} was derived from interaction records between counselors and clients on real online counseling platforms.

On the other hand, LLM-augmented datasets were created by having LLMs simulate both the counselor and client roles \cite{chen-etal-2023-soulchat, SweetieChat}. For instance, HealMe \cite{xiao-etal-2024-healme} and C\scalebox{0.8}{ACTUS} \cite{lee-etal-2024-cactus} generated dialogues based on cognitive behavioral therapy (CBT) using carefully designed prompts; ESD-CoT \cite{zhang-etal-2024-escot} extracted  situations from existing datasets to generate complete dialogues. SMILECHAT \cite{qiu-etal-2024-smile} expanded single-turn Q\&A into multi-turn dialogues, while A\scalebox{0.8}{UG}ESC \cite{zheng-etal-2023-augesc} modeled data augmentation as a dialogue completion task to extend conversations.
As shown in Table~\ref{tab:dataset_comparison}, although these datasets are often large in size, they tend to contain fewer utterances and continue to face challenges in content diversity \cite{zheng-etal-2024-self}.
In contrast, manually constructed datasets are smaller while offering higher authenticity and quality. This study proposes the largest known human-collected psychological counseling dialogue dataset, with dialogue lengths resembling real counseling sessions.

\subsection{Evaluation of Counseling Dialogue}

In recent years, studies have increasingly explored the use of LLMs for dialogue evaluation in specific scenarios. For example, \citet{ChatCounselor} employed GPT-4 to compare psychological counseling responses generated by different models across multiple dimensions, such as information quality and user self-disclosure. \citet{lee-etal-2024-cactus} and \citet{zhao-etal-2024-esc} simulated client interactions with counselor models, conducting full-length dialogues and evaluating them based on overall performance.

Additionally, some datasets have introduced rating mechanisms. For instance, the Client-Reactions \cite{li-etal-2023-understanding} recorded client ratings across four dimensions alongside dialogue collection. However, among 2,382 dialogue turns, only 479 turns included ratings, limiting its scale.
In contrast, our study introduces a dataset with 6,589 counseling dialogues, all accompanied by rating information. We design 20 evaluation dimensions (as shown in Table~\ref{tab:feedback}) to assess both the overall impression of the counseling session and the counselor’s professional skills, providing a more comprehensive standard for evaluating dialogue quality.

\section{Data Collection}

To construct a high-quality psychological counseling dialogue dataset, we employed a role-playing approach in which trained professional and trainee counselors simulated counselor-client interactions, ensuring authenticity and professional relevance. Additionally, we collected detailed client feedback, providing a valuable resource for evaluating psychological counseling dialogues.

\subsection{Data Source}
\label{sec:data_source}

We developed an online platform to facilitate participant matching, role-playing dialogues, and client feedback collection. On this platform, participants can choose their preferred roles and schedule dialogues based on their availability and role preferences. Once matched, role-playing dialogues take place at the designated time, with the counselor-role player communicating via a computer keyboard and the client-role player using LINE, a mobile messaging app. This setup reflects real-world online text-based psychological counseling practices in Japan.
Additionally, to ensure a complete record of the dialogue process, the platform stores message timestamps for further analysis. The detailed interface is shown in Appendix~\ref{sec:online_platform}.


Each dialogue session typically lasts one hour, though the duration may be adjusted as needed. As part of the role-playing process, the counselor-role player can specify discussion topics and the client’s psychological state (e.g., whether suicidal tendencies are present). If no specific conditions are set, the client-role player decides freely.
To protect participant privacy, all client-role participants are explicitly instructed not to discuss their real-life concerns during the dialogue and are strictly prohibited from sharing personal identifying information. After the session, the client-role participant evaluates the counselor-role player’s performance, with specific evaluation criteria detailed in Section~\ref{sec:client_feedback}.

\subsection{Participants}

The dataset comprises 480 participants, including 117 males, 360 females, and 3 individuals who did not disclose their gender. Participants' ages range from 21 to 78, with approximately 80\% between 30 and 59 years old. Detailed distribution information is provided in Appendix~\ref{sec:age_distribution}.
All participants are native Japanese speakers, with over 80\% having played both the client and counselor roles. In total, 424 participants took on the counselor role, and 463 participants played the client role.

\paragraph{Professionalism} 

Participants have expertise in online psychological counseling. More than one-third hold professional qualifications and have practical counseling experience, while the rest, though not yet certified, have undergone six months to one year of systematic study with the goal of obtaining certification. Additionally, all participants completed a 10-hour structured training program covering the characteristics, advantages, and limitations of online text-based psychological counseling, the role and ethical guidelines of counselors, as well as professional counseling skills and procedures.

\subsection{Client Feedback}
\label{sec:client_feedback}

\begin{table*}[t!]
\centering
\footnotesize
\renewcommand{\arraystretch}{0.5}  
\setlength{\tabcolsep}{6pt}  
\begin{tabular}{m{2.7cm} m{3cm} m{6cm}} 
\toprule
  \textbf{Category} & \textbf{Aspect} & \textbf{Feedback Item} \\ 
\midrule

  \multirow{19}{=}{\makecell[l]{\textbf{Overall Impression} \\ \textbf{of the Conversation} }} 
  & \textbf{Sense of Validation} 
    & 1. Felt heard and understood. \newline 2. Felt respected. \\ \cmidrule(l){2-3}
  & \textbf{Awareness and Hope} 
    & 3. Gained new insights. \newline 4. Felt hopeful or expectant. \\ \cmidrule(l){2-3}
  & \textbf{Engagement} 
    & 5. Concerns were addressed. \newline 6. Thought through concerns together. \\ \cmidrule(l){2-3}
  & \textbf{Flow and Comfort} 
    & 7. The conversation had a good rhythm. \newline 8. The conversation felt comfortable. \\ \cmidrule(l){2-3}
  & \textbf{Overall Evaluation} 
    & 9. Felt appropriate and satisfying. \newline 10. The conversation was valuable. \\ 
\midrule

  \multirow{19}{=}{\makecell[l]{\textbf{Evaluation of} \\ \textbf{Counseling Skills}}} 
  & \textbf{Flow of Conversation} 
    & 11. The conversation started smoothly. \newline 12. The conversation ended well. \\ \cmidrule(l){2-3}
  & \textbf{Counseling Skills} 
    & 13. Showed acceptance and empathy. \newline
      14. Provided acknowledgment and affirmation. \newline
      15. Asked effective questions to foster dialogue. \newline
      16. Summarized key points effectively. \newline
      17. Clarified issues clearly. \newline
      18. Helped identify goals for the conversation. \newline
      19. Offered actionable suggestions. \newline
      20. Encouraged and instilled hope. \\ 
\bottomrule
\end{tabular}
\caption{The 20 client feedback items (each rated on a 0–5 point scale).}
\label{tab:feedback}
\end{table*}


After each role-playing, the client-role player evaluates the counselor-role player’s performance. The results are immediately shared with the counselor-role player and are monitored by the platform administrator to ensure fairness and reliability.

The client feedback items are designed under the supervision of an expert holding the nationally recognized Certified Public Psychologist qualification and a Ph.D. degree. As shown in Table~\ref{tab:feedback}, the feedback covers two main aspects:
\begin{itemize}
\item \textbf{Overall impression of the conversation} (e.g., understanding and respect, sense of hope, engagement, fluency, satisfaction)
\item \textbf{Evaluation of counseling skills} (e.g., empathy, affirmation, effective questioning, goal setting, problem clarification, and conveying hope)
\end{itemize}

The evaluation employs a six-point Likert scale (0–5 points) across 20 items, with a maximum total score of 100 points. Additionally, three check items assess serious issues, such as inappropriate remarks or ethical violations. If any of these are selected, the overall score is halved or reset to zero. Further details are provided in Appendix~\ref{sec:screen_item}.

\section{Data Characteristics}

This study collected dialogue data from March 7, 2020, to September 8, 2024, filtering out conversations with fewer than 30 utterances, durations under 30 minutes, or cases where all 20 evaluation items were rated as 3 (as such scores may be unreliable). The final dataset consists of 6,589 dialogues, with statistical details provided in Table~\ref{tab:conversation_stats}.

Each dialogue contains an average of 91.20 utterances, surpassing other manually collected datasets such as ESConv \cite{liu-etal-2021-towards} (29.5 utterances), Anno-MI \cite{AnnoMi} (72.9 utterances), and Client-Reactions \cite{li-etal-2023-understanding} (78.5 utterances). This suggests that our dataset provides greater depth and interactivity. Additionally, the average utterance length of counselors (35.84 characters) is significantly higher than that of clients (20.63 characters), reflecting the counselor’s guiding role in conversations, where they typically use more detailed language to provide support.

Additionally, the dataset includes 480 unique speakers, with 424 counselors and 463 clients, resulting in 4,900 distinct counselor–client pairings, which contribute to conversational diversity to some extent.

\begin{table}[t!]
\renewcommand{\arraystretch}{1.4} 
\small
\centering
\setlength{\tabcolsep}{1pt} 
\begin{tabular}{l c c c}
\toprule
\textbf{Category} & \textbf{Total} & \textbf{Counselor} & \textbf{Client} \\
\midrule
\textbf{\# Dialogues} & 6,589 & - & - \\  
\textbf{\# Speakers} & 480 & 424 & 463 \\
\textbf{\# Utterances} & 600,939 & 306,495 & 294,444 \\
\textbf{Avg. utterances per dialogue} & 91.20 & 46.52 & 44.69 \\
\textbf{Avg. length per utterance} & 28.39 & 35.84 & 20.63 \\
\bottomrule
\end{tabular}
\caption{Statistics of the overall conversations. An "utterance" denotes one discrete message sent by a client or counselor in the chat system upon clicking the send button.}
\label{tab:conversation_stats}
\end{table}

\subsection{Dialogue Topics}

The counseling topics were determined by the participants. To gain a comprehensive understanding of the issues present in KokoroChat, we utilized GPT-4o-mini \cite{gpt4o} to analyze dialogue topics. Specifically, we input the dialogue content to predict problem types and generate more detailed descriptions. The prompt used for this analysis is provided in Appendix~\ref{sec:prompt_topic}.

\begin{figure}[t!]
  \centering
  \includegraphics[width=0.99\linewidth]{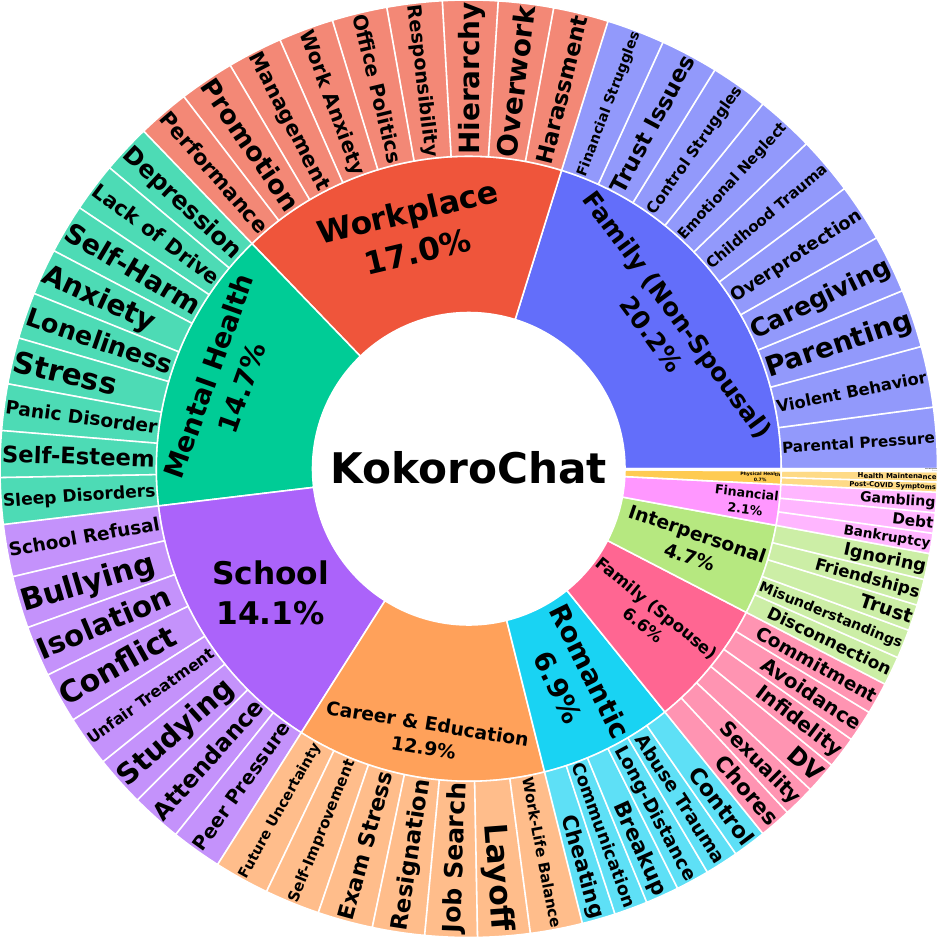}
  \caption{The distribution of issues in KokoroChat.}
  \label{fig:fig_topic_chart}
\end{figure}


The distribution of the 11 predefined issue types is shown in Figure~\ref{fig:fig_topic_chart}. Family issues (non-spousal) account for the largest proportion (20.2\%), followed by workplace issues (17.0\%) and mental health issues (14.7\%), reflecting that family, work, and mental health are primary concerns for clients. 
School issues (14.1\%) and career and education issues (12.9\%) also constitute significant portions, highlighting the importance of education and career development in counseling conversations.

Based on the generated detailed descriptions, we further summarized the characteristics of each issue category, as presented in Figure~\ref{fig:fig_topic_chart}. Overall, the dataset encompasses a broad range of real-world problems, demonstrating high diversity and providing a solid foundation for research on psychological counseling dialogues.


\subsection{Analysis on Client Feedback}
\label{sec:data_analysis}

\paragraph{Score Distribution} 

\begin{figure}[t!]
  \centering
  \includegraphics[width=\linewidth]{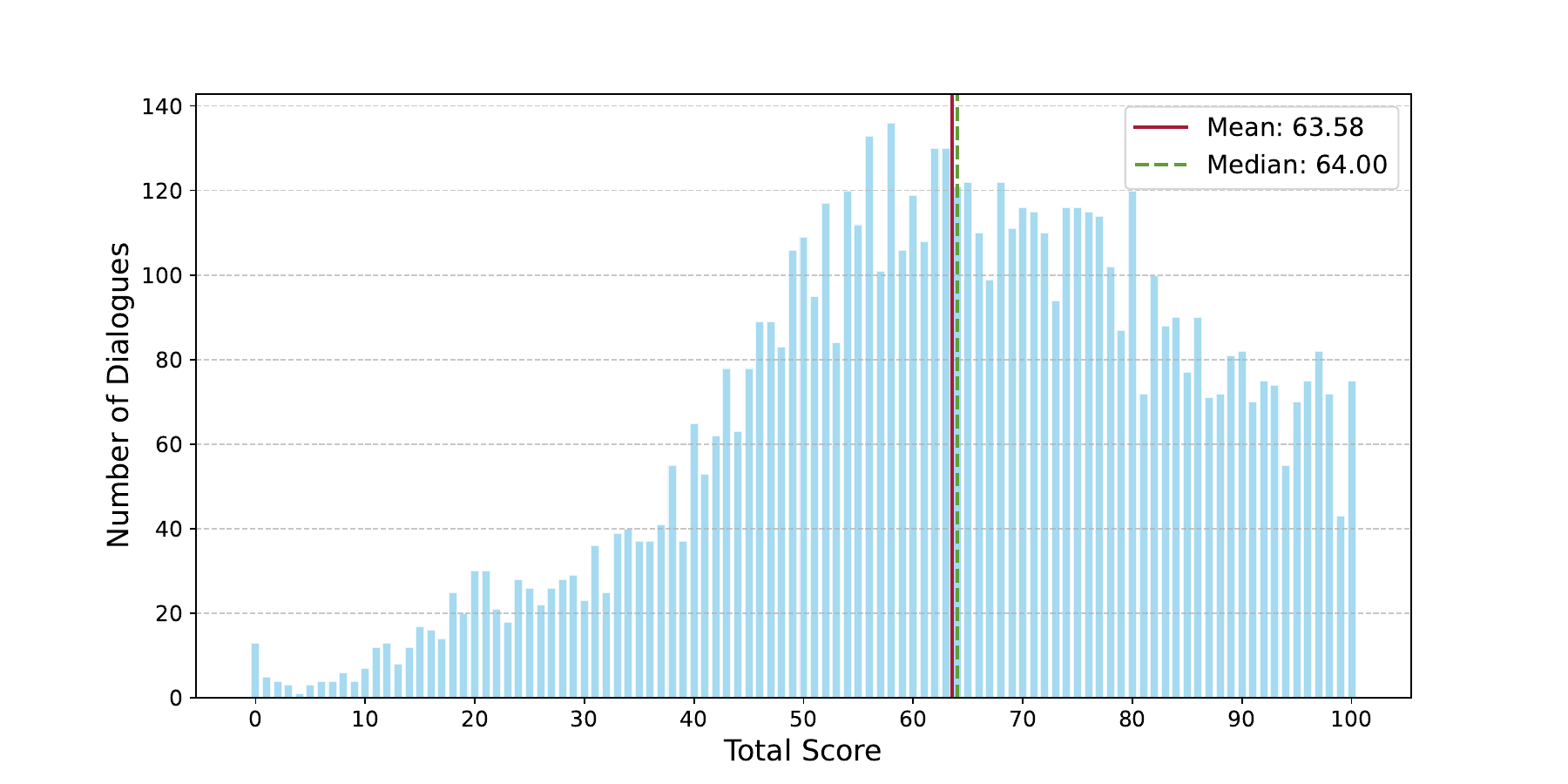}
  \caption{Score distribution of dialogues.}
  \label{fig:fig_score_distribution}
\end{figure}

Figure~\ref{fig:fig_score_distribution} presents the score distribution of dialogues in KokoroChat. The histogram shows a well-balanced distribution, with a unimodal shape centered around the mean (63.58) and median (64.00). Additionally, the distribution exhibits a slight right skew, indicating that most dialogues received moderate to high client feedback.

\paragraph{Correlation Between Dialogue Features and Scores}

\begin{figure}[t!]
  \centering
  \includegraphics[width=\linewidth]{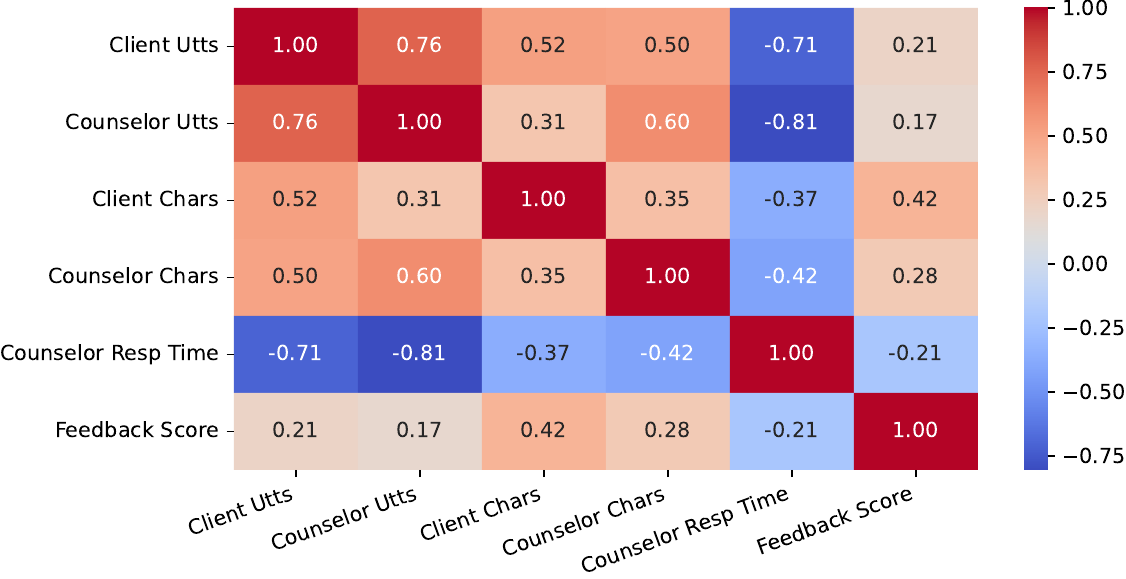}
  \caption{Spearman correlation between dialogue features and feedback scores.}
  \label{fig:fig_spearman_correlation_main}
\end{figure}

We conducted a Spearman correlation analysis to examine the relationship between various dialogue features and feedback scores, as shown in Figure~\ref{fig:fig_spearman_correlation_main}. The results indicate that the total word count of the client has the highest positive correlation with the score ($\rho = 0.42$), suggesting that the extent of client expression may influence their evaluation. When clients use more words to express themselves, they may feel better heard and understood, leading to a more positive assessment of their counseling experience.
In contrast, the total word count of the counselor shows a lower correlation with the score ($\rho = 0.28$), implying that while greater counselor speech may contribute to higher ratings, its impact is relatively limited.
Additionally, the correlation between utterance count and scores is weaker, with clients ($\rho = 0.21$) and counselors ($\rho = 0.17$) both showing a positive correlation, though to a lesser extent than word count.
This result suggests that the richness of conveyed information may be more influential than the number of utterances.
Furthermore, the counselor’s average response time exhibits a negative correlation with the scores ($\rho = -0.21$), indicating that longer response times may negatively impact user experience, whereas quicker responses could potentially contribute to higher ratings.


\paragraph{Correlation Among Evaluation Dimensions}

Additionally, we conducted a Spearman correlation analysis to examine the relationships between different evaluation dimensions. The results indicate that most rating dimensions exhibit a strong positive correlation ($\rho > 0.6$), suggesting that the counseling experience is influenced by multiple interrelated factors rather than a single determinant.

Notably, D1 (\textit{felt heard and understood}) shows a high correlation with D2 (\textit{felt respected}), D9 (\textit{felt appropriate and satisfying}), and D10 (\textit{the conversation was valuable}). This suggests that when clients feel heard and understood, they are more likely to experience a sense of respect, perceive the counseling process as meaningful, and ultimately report higher overall satisfaction.
The complete results are presented in Figure ~\ref{fig:fig_spearman_correlation_matrix} in Appendix~\ref{sec:client_feedback_analysis}.

\section{Experiments}

To evaluate KokoroChat's potential in psychological counseling response generation and dialogue assessment, we conducted experiments on two tasks: response generation and score prediction.

\subsection{Response Generation}

Due to the lack of a Japanese psychological counseling dataset and a Japanese LLM specifically designed for counseling, direct comparison with other models is not feasible\footnote{To provide a rough performance reference, we conducted a simplified experiment comparing our model with models trained on non-Japanese counseling data in a Japanese setting. See Appendix~\ref{sec:add_experiment} for details.}. Therefore, this study focuses on verifying whether fine-tuning on KokoroChat can enhance the performance of open-source LLMs in psychological counseling tasks.

For dialogue data preprocessing, we applied an utterance merging strategy, combining consecutive utterances from the same speaker into a single utterance. The model takes the complete dialogue history \( D_t = \{ u_1^C, u_2^S, u_3^C, \dots, u_t^C \} \) as input, where \( u_i^C \) and \( u_j^S \) represent utterances from the client (\( C \)) and counselor (\( S \)), respectively. The model then generates the next counselor response \( u_{t+1}^S \).

\subsubsection{Models}

In this experiment, we used Llama 3.1 Swallow \footnote{\href{https://huggingface.co/tokyotech-llm/Llama-3.1-Swallow-8B-Instruct-v0.3}{Llama-3.1-Swallow-8B-Instruct-v0.3} was used, which is a continuously pre-trained variant of Meta Llama 3.1 \cite{dubey2024llama3herdmodels} with enhanced Japanese proficiency and optimized dialogue generation.} \cite{Fujii:COLM2024, Okazaki:COLM2024} as the base model, fine-tuning and evaluating it using the KokoroChat dataset. To ensure high-quality test data, we selected 118 dialogues with client feedback scores of 99 and 100 as the test set, while the remaining data was used for fine-tuning. 
The model-generated responses were then compared with the corresponding counselor replies in the test set to evaluate the quality of the generated outputs.

To explore the impact of client feedback on model ability to generate psychological counseling responses, we constructed the following variants:
\begin{itemize}
\item \textbf{Kokoro-Low:} Fine-tuned on 334,022 utterances from 3,870 dialogues with a client feedback score of $< 70$.\footnote{Considering that low-scoring dialogues tend to have a relatively lower number of utterances, we set the threshold at 70 to ensure balanced data segmentation.} 
\item \textbf{Kokoro-High:} Fine-tuned on 254,515 utterances from 2,601 dialogues with a client feedback score in the range of $70 \leq score \leq 98$.
\item \textbf{Kokoro-Full:} Fine-tuned on 6,471 dialogues with a client feedback score of $\leq 98$.
\end{itemize}

Finally, we evaluated these variants on the test set to examine the impact of data partitions on model improvement. 
Additionally, we compared our models with GPT-4o \cite{gpt4o}, one of the most advanced models, to further understand their relative performance.
Appendix~\ref{sec:model_train} provides details on the training process.

\subsubsection{Automatic Evaluation}
We used three commonly adopted automatic evaluation metrics: BLEU-n \cite{bleu}, ROUGE-L \cite{lin-2004-rouge}, and Distinct-n \cite{li-etal-2016-diversity}. The results are presented in Table~\ref{tab:performance_comparison}. Experimental findings indicate that Kokoro-High performs best on most BLEU and ROUGE metrics, likely due to the higher quality of its training data and its closer alignment with the test set. In contrast, Kokoro-Full, which includes a larger dataset, achieves slightly better performance on the diversity metric Dist-n. For non-fine-tuned models, GPT-4o outperformed Llama-3.1 across all automatic evaluation metrics.

\begin{table*}[t]
    \centering
    \small
    \renewcommand{\arraystretch}{1.4}
    \setlength{\tabcolsep}{4pt}
    \begin{tabular}{lccccccccc}
        \toprule
        \textbf{Model} & \textbf{BLEU-1} & \textbf{BLEU-2} & \textbf{BLEU-3} & \textbf{BLEU-4} & \textbf{ROUGE-1} & \textbf{ROUGE-2} & \textbf{ROUGE-L} & \textbf{Dist-1} & \textbf{Dist-2} \\
        \midrule
        Llama-3.1    & 17.32 & 9.13  & 4.77  & 2.25  & 23.81 & 7.37  & 16.96 & 1.04  & 6.86  \\
        GPT-4o  & 21.77 & 11.72  & 6.32  & 3.17  & 28.67 & 9.19  & 19.82 & 1.19  & 6.90  \\
        Kokoro-Low   & 25.39 & 15.30 & 8.69  & 5.39  & 33.38 & 14.05 & 27.28 & \underline{2.42}  & 12.98 \\
        Kokoro-High  & \textbf{27.03} & \textbf{16.45} & \textbf{9.57}  & \textbf{6.00}  & \textbf{34.64} & \textbf{14.72} & \underline{28.00} & 2.33  & \underline{13.08} \\
        Kokoro-Full  & \underline{25.69} & \underline{15.65} & \underline{9.23}  & \underline{5.83}  & \underline{34.02} & \underline{14.60} & \textbf{28.10} & \textbf{2.48}  & \textbf{13.24} \\
        \bottomrule
    \end{tabular}
    \caption{Performance comparison of models. Best values are in \textbf{bold}, second-best are \underline{underlined}.}
    \label{tab:performance_comparison}
\end{table*}

\subsubsection{Human Evaluation}

We also conducted a human evaluation of 100 responses generated by each model, independently assessed by five professional counselors. Specifically, we randomly selected 10 dialogues from the test set and, for each dialogue, randomly sampled 10 sets of dialogue histories of varying lengths to generate model responses. 
The evaluation used pairwise comparison, where counselors judged which response was more suitable (Win, Lose, Tie). The final result followed majority voting—if over half agreed, it was adopted; otherwise, or if Tie votes exceeded half, the result was Tie.

\begin{figure}[t!]
  \centering
  \includegraphics[width=\linewidth]{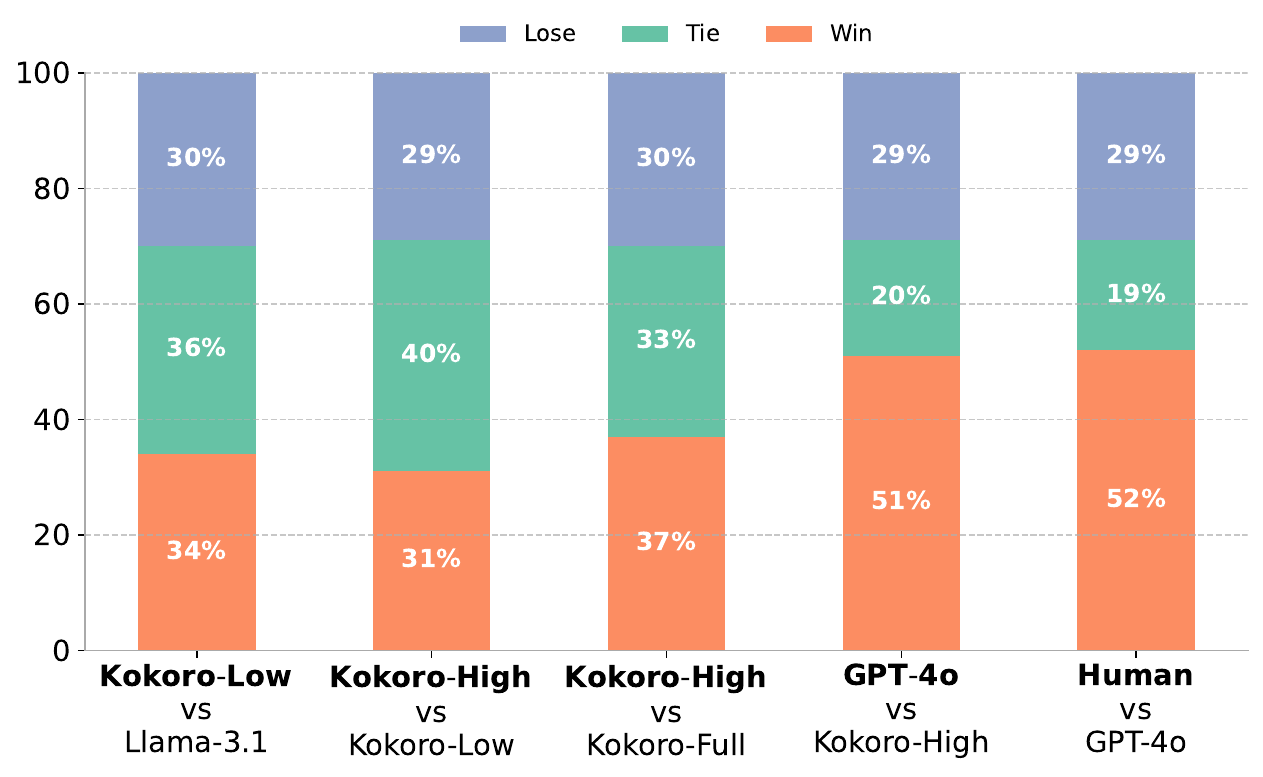}
  \caption{Human evaluation result. Orange denotes upper model wins; the winning model is in \textbf{bold}. \textit{Human} refers to responses from human counselors in the KokoroChat test dataset.}
  \label{fig:fig_human_eval}
\end{figure}

Figure~\ref{fig:fig_human_eval} presents the evaluation results. The comparison between Kokoro-Low and Llama-3.1 indicates that even when using only the lower-scoring portions of KokoroChat, it still enhances open-source LLMs in generating psychological counseling responses. 
Notably, despite using less data, Kokoro-High outperforms both Kokoro-Low and Kokoro-Full, similar to the results of automatic evaluation, highlighting the importance of high-quality training data in improving model performance.
However, due to the difference in model size (Llama-3.1 = 8B, GPT-4o \( \approx \) 200B\footnote{This is merely an estimate by \citet{medec}.}), the fine-tuned model still lags behind GPT-4o. Similarly, GPT-4o's responses also exhibit a noticeable gap compared to those of highly rated human counselors, further emphasizing the high quality of KokoroChat. The generated response examples are shown in Figure~\ref{fig:fig_case_study} of Appendix~\ref{sec:case_study}.

\subsection{Score Prediction}
Automatic evaluation of psychological counseling dialogues not only reduces the cost of human evaluation but also provides counselors with feedback to enhance their professional skills. To verify the effectiveness of LLMs fine-tuned on KokoroChat for dialogue evaluation, we conducted a score prediction experiment. 

Specifically, given a complete dialogue \( D \) as input, the model predicts a set of item-score pairs \( S = \{ (d_1, s_1), (d_2, s_2), \dots, (d_{20}, s_{20}) \} \), where each \( d_i \) (\( i = 1, 2, \dots, 20 \)) represents an evaluation dimension (e.g., \textit{felt heard and understood}, \textit{felt respected}), and each \( s_i \) corresponds to a score in the range \( [0,5] \).




\subsubsection{Models}

We similarly employed Llama-3.1-Swallow as the base model and fine-tuned it by splitting the dataset into training, validation, and test sets (8:1:1). To ensure robust results, we repeated fine-tuning five times with different seeds. For comparison, we also performed zero-shot score prediction using the original Llama-3.1-Swallow and GPT-4o.
Detailed training procedures are provided in Appendix~\ref{sec:model_train}.

\subsubsection{Results}

\begin{table}[t]
    \centering
    \small
    \renewcommand{\arraystretch}{1.4}
    \setlength{\tabcolsep}{4pt}
    \begin{tabular}{lccc}
        \toprule
        \textbf{Model} & \textbf{ACC} (\textbf{\textuparrow}) & $\mathbf{ACC_{soft}}$ (\textbf{\textuparrow}) & \textbf{MAE} (\textbf{\textdownarrow}) \\
        \midrule
        Llama-3.1  & 28.70 \scriptsize{$\pm$ 7.39} & 72.53 \scriptsize{$\pm$ 12.40} & 1.0540 \scriptsize{$\pm$ 0.2731} \\
        GPT-4o     & 30.92 \scriptsize{$\pm$ 6.84} & 75.27 \scriptsize{$\pm$ 11.04} & 1.0151 \scriptsize{$\pm$ 0.2685} \\
        \rowcolor{gray!15} \textbf{Ours}  & \textbf{35.35} \scriptsize{$\pm$ \textbf{1.75}} & \textbf{83.64} \scriptsize{$\pm$ \textbf{2.15}} & \textbf{0.8283} \scriptsize{$\pm$ \textbf{0.0349}} \\
        \bottomrule
    \end{tabular}
    \caption{Performance comparison of different models in terms of accuracy (ACC), soft accuracy ($\mathrm{ACC_{soft}}$), and mean absolute error (MAE) (the results of model \textbf{Ours} are averaged over five different random seeds; detailed results can be found in Table~\ref{tab:performance_models_5seeds} of Appendix~\ref{sec:model_train}).}
    \label{tab:performance_score_pre}
\end{table}


Table~\ref{tab:performance_score_pre} presents the average performance of different models in predicting scores across 20 evaluation dimensions for psychological counseling dialogues.
Our model outperforms Llama-3.1 and GPT-4o in accuracy, demonstrating superior score prediction capabilities. Given the inherent subjectivity and ambiguity of human ratings, we also evaluated performance using $\mathrm{ACC_{soft}}$, which allows a ±1 error margin between predicted and human-assigned scores. Our model again surpasses the baselines under this metric, highlighting its ability to capture scoring trends in psychological counseling dialogues while maintaining robust performance with a more flexible scoring standard. Detailed results across 20 dimensions are in Table~\ref{tab:performance_D1_D20} of Appendix~\ref{sec:model_train}.

Our model also achieves the lowest Mean Absolute Error (MAE), indicating smaller prediction errors and closer alignment with human ratings. Additionally, its lower standard deviation across all metrics suggests more stable scoring. These results confirm the effectiveness of the KokoroChat dataset and demonstrate that our fine-tuned model delivers both stable and high-performance predictions across 20 evaluation dimensions.



\section{Conclusion}

This study introduces \textbf{KokoroChat}, the largest manually collected psychological counseling dialogue dataset to date, developed using a role-playing approach. The dataset includes detailed client feedback, enabling automatic evaluation of psychological counseling dialogues.
Experimental results demonstrate that KokoroChat enhances LLM performance in generating psychological counseling responses. Additionally, by leveraging its extensive client feedback data, we train a dialogue evaluation model capable of producing more robust and accurate assessment results.

\section*{Limitations}

Due to the lack of publicly available Japanese psychological counseling datasets or Japanese LLMs designed for counseling, this study could not be directly compared with existing research. As part of future work, we plan to translate KokoroChat into multiple languages, such as Chinese and English, enabling comparisons with other psychological counseling dialogue datasets and models for a more comprehensive evaluation of our approach.

Additionally, we plan to annotate dialogue acts within the dataset to analyze the evolving strategies used in counseling and their impact on outcomes. This will provide deeper insights into how counselors adjust their communication styles based on client responses and help optimize the model’s adaptability across different counseling scenarios.

Furthermore, potential gender and age biases among participants during data collection may affect the model's generalization ability. 

\section*{Ethical Considerations}

The dialogue data collected in this study originates from an internal training platform used by a psychological counselor association. This platform is designed to support professional counselors and aspiring trainees in developing their psychological counseling skills.
Psychological counseling relies not only on a strong theoretical foundation but also on extensive practical experience. Even experienced counselors must engage in continuous practice to refine their skills. However, gaining experience in real counseling settings can pose ethical and safety risks, particularly when clients are experiencing emotional distress or psychological crises. Counselors cannot rely solely on real-world counseling experiences for training. To address this, the platform provides a low-risk training environment, enabling counselors to practice and refine their skills in simulated scenarios.

The dialogues on this platform are not real psychological counseling sessions; rather, they are role-play-based simulated counseling exercises. All participants are fully aware of the simulated nature of the dialogues and voluntarily engage in the training process without monetary compensation. As a benefit, participants gain valuable hands-on experience and receive feedback from role-players with relevant professional backgrounds, helping them further develop their counseling skills.

Due to the nature of psychological counseling, even in role-play scenarios, dialogues may include expressions of severe emotional distress, such as suicidal ideation or other extreme emotions. These simulated cases are designed to help counselors develop crisis intervention skills in a controlled setting. While these dialogues do not represent real client experiences, they reflect situations that counselors may encounter in real practice, contributing to their preparedness for handling complex emotional states.

Additionally, to protect participant privacy, all users are explicitly informed that they must not discuss real-life personal issues during the dialogues and must not disclose real names or other identifiable information. 
Furthermore, they are fully aware that their dialogue data will be stored on the training platform and may be used for service optimization, scientific research, or third-party academic studies in the future.

\section*{Acknowledgments}
The authors wish to acknowledge Prof. Yasushi Sugihara from Kyoto University for his insightful advice during the design of the client feedback items.
This work was supported by JSPS KAKENHI Grant Number 25H01382. 
\bibliography{custom}

\begin{thebibliography}{31}
\providecommand{\natexlab}[1]{#1}

\bibitem[{Abacha et~al.(2024)Abacha, wai Yim, Fu, Sun, Yetisgen, Xia, and Lin}]{medec}
Asma~Ben Abacha, Wen wai Yim, Yujuan Fu, Zhaoyi Sun, Meliha Yetisgen, Fei Xia, and Thomas Lin. 2024.
\newblock \href {https://arxiv.org/abs/arXiv:2412.19260} {Medec: A benchmark for medical error detection and correction in clinical notes}.

\bibitem[{Althoff et~al.(2016)Althoff, Clark, and Leskovec}]{skillful}
Tim Althoff, Kevin Clark, and Jure Leskovec. 2016.
\newblock \href {https://doi.org/10.1162/tacl_a_00111} {Large-scale analysis of counseling conversations: An application of natural language processing to mental health}.
\newblock \emph{Transactions of the Association for Computational Linguistics}, 4:463--476.

\bibitem[{Chen et~al.(2023)Chen, Xing, Lin, Zheng, Wang, Liu, and Xu}]{chen-etal-2023-soulchat}
Yirong Chen, Xiaofen Xing, Jingkai Lin, Huimin Zheng, Zhenyu Wang, Qi~Liu, and Xiangmin Xu. 2023.
\newblock \href {https://doi.org/10.18653/v1/2023.findings-emnlp.83} {{S}oul{C}hat: Improving {LLM}s' empathy, listening, and comfort abilities through fine-tuning with multi-turn empathy conversations}.
\newblock In \emph{Findings of the Association for Computational Linguistics: EMNLP 2023}, pages 1170--1183, Singapore. Association for Computational Linguistics.

\bibitem[{Dettmers et~al.(2023)Dettmers, Pagnoni, Holtzman, and Zettlemoyer}]{qlora}
Tim Dettmers, Artidoro Pagnoni, Ari Holtzman, and Luke Zettlemoyer. 2023.
\newblock \href {https://arxiv.org/abs/arXiv:2305.14314} {Qlora: Efficient finetuning of quantized llms}.

\bibitem[{Dubey et~al.(2024)Dubey, Jauhri, Pandey, Kadian, Al-Dahle, Letman, Mathur, Schelten, Yang, and et~al.}]{dubey2024llama3herdmodels}
Abhimanyu Dubey, Abhinav Jauhri, Abhinav Pandey, Abhishek Kadian, Ahmad Al-Dahle, Aiesha Letman, Akhil Mathur, Alan Schelten, Amy Yang, and Angela~Fan et~al. 2024.
\newblock \href {https://arxiv.org/abs/2407.21783} {The llama 3 herd of models}.
\newblock \emph{Preprint}, arXiv:2407.21783.

\bibitem[{Fujii et~al.(2024)Fujii, Nakamura, Loem, Iida, Ohi, Hattori, Shota, Mizuki, Yokota, and Okazaki}]{Fujii:COLM2024}
Kazuki Fujii, Taishi Nakamura, Mengsay Loem, Hiroki Iida, Masanari Ohi, Kakeru Hattori, Hirai Shota, Sakae Mizuki, Rio Yokota, and Naoaki Okazaki. 2024.
\newblock Continual pre-training for cross-lingual llm adaptation: Enhancing japanese language capabilities.
\newblock In \emph{Proceedings of the First Conference on Language Modeling}, COLM, University of Pennsylvania, USA.

\bibitem[{Inaba et~al.(2024)Inaba, Ukiyo, and Takamizo}]{inaba2024}
Michimasa Inaba, Mariko Ukiyo, and Keiko Takamizo. 2024.
\newblock \href {https://arxiv.org/abs/2402.12738} {Can large language models be used to provide psychological counselling? an analysis of gpt-4-generated responses using role-play dialogues}.
\newblock In \emph{The 14th International Workshop on Spoken Dialogue Systems Technology}.

\bibitem[{Lee et~al.(2024)Lee, Kim, Kim, Kang, Yang, Kim, Kang, Jung, Kim, Lee, Chung, Yu, Lee, and Yeo}]{lee-etal-2024-cactus}
Suyeon Lee, Sunghwan Kim, Minju Kim, Dongjin Kang, Dongil Yang, Harim Kim, Minseok Kang, Dayi Jung, Min~Hee Kim, Seungbeen Lee, Kyong-Mee Chung, Youngjae Yu, Dongha Lee, and Jinyoung Yeo. 2024.
\newblock \href {https://doi.org/10.18653/v1/2024.findings-emnlp.832} {Cactus: Towards psychological counseling conversations using cognitive behavioral theory}.
\newblock In \emph{Findings of the Association for Computational Linguistics: EMNLP 2024}, pages 14245--14274, Miami, Florida, USA. Association for Computational Linguistics.

\bibitem[{Li et~al.(2023)Li, Ma, Mei, He, Zhang, Qiu, and Lan}]{li-etal-2023-understanding}
Anqi Li, Lizhi Ma, Yaling Mei, Hongliang He, Shuai Zhang, Huachuan Qiu, and Zhenzhong Lan. 2023.
\newblock \href {https://doi.org/10.18653/v1/2023.acl-long.577} {Understanding client reactions in online mental health counseling}.
\newblock In \emph{Proceedings of the 61st Annual Meeting of the Association for Computational Linguistics (Volume 1: Long Papers)}, pages 10358--10376, Toronto, Canada. Association for Computational Linguistics.

\bibitem[{Li et~al.(2016)Li, Galley, Brockett, Gao, and Dolan}]{li-etal-2016-diversity}
Jiwei Li, Michel Galley, Chris Brockett, Jianfeng Gao, and Bill Dolan. 2016.
\newblock \href {https://doi.org/10.18653/v1/N16-1014} {A diversity-promoting objective function for neural conversation models}.
\newblock In \emph{Proceedings of the 2016 Conference of the North {A}merican Chapter of the Association for Computational Linguistics: Human Language Technologies}, pages 110--119, San Diego, California. Association for Computational Linguistics.

\bibitem[{Lin(2004)}]{lin-2004-rouge}
Chin-Yew Lin. 2004.
\newblock \href {https://aclanthology.org/W04-1013/} {{ROUGE}: A package for automatic evaluation of summaries}.
\newblock In \emph{Text Summarization Branches Out}, pages 74--81, Barcelona, Spain. Association for Computational Linguistics.

\bibitem[{Liu et~al.(2023)Liu, Li, Cao, Ren, Liao, and Wu}]{ChatCounselor}
June~M. Liu, Donghao Li, He~Cao, Tianhe Ren, Zeyi Liao, and Jiamin Wu. 2023.
\newblock \href {https://arxiv.org/abs/arXiv:2309.15461} {Chatcounselor: A large language models for mental health support}.

\bibitem[{Liu et~al.(2021)Liu, Zheng, Demasi, Sabour, Li, Yu, Jiang, and Huang}]{liu-etal-2021-towards}
Siyang Liu, Chujie Zheng, Orianna Demasi, Sahand Sabour, Yu~Li, Zhou Yu, Yong Jiang, and Minlie Huang. 2021.
\newblock \href {https://doi.org/10.18653/v1/2021.acl-long.269} {Towards emotional support dialog systems}.
\newblock In \emph{Proceedings of the 59th Annual Meeting of the Association for Computational Linguistics and the 11th International Joint Conference on Natural Language Processing (Volume 1: Long Papers)}, pages 3469--3483, Online. Association for Computational Linguistics.

\bibitem[{Okazaki et~al.(2024)Okazaki, Hattori, Shota, Iida, Ohi, Fujii, Nakamura, Loem, Yokota, and Mizuki}]{Okazaki:COLM2024}
Naoaki Okazaki, Kakeru Hattori, Hirai Shota, Hiroki Iida, Masanari Ohi, Kazuki Fujii, Taishi Nakamura, Mengsay Loem, Rio Yokota, and Sakae Mizuki. 2024.
\newblock Building a large japanese web corpus for large language models.
\newblock In \emph{Proceedings of the First Conference on Language Modeling}, COLM, University of Pennsylvania, USA.

\bibitem[{OpenAI(2023)}]{gpt4}
OpenAI. 2023.
\newblock \href {https://arxiv.org/abs/arXiv:2303.08774} {Gpt-4 technical report}.

\bibitem[{OpenAI(2024)}]{gpt4o}
OpenAI. 2024.
\newblock \href {https://arxiv.org/abs/arXiv:2410.21276} {Gpt-4o system card}.

\bibitem[{Papineni et~al.(2002)Papineni, Roukos, Ward, and Zhu}]{bleu}
Kishore Papineni, Salim Roukos, Todd Ward, and Wei-Jing Zhu. 2002.
\newblock \href {https://doi.org/10.3115/1073083.1073135} {Bleu: a method for automatic evaluation of machine translation}.
\newblock In \emph{Proceedings of the 40th Annual Meeting on Association for Computational Linguistics}, ACL '02, page 311–318, USA. Association for Computational Linguistics.

\bibitem[{Qiu et~al.(2024)Qiu, He, Zhang, Li, and Lan}]{qiu-etal-2024-smile}
Huachuan Qiu, Hongliang He, Shuai Zhang, Anqi Li, and Zhenzhong Lan. 2024.
\newblock \href {https://doi.org/10.18653/v1/2024.findings-emnlp.34} {{SMILE}: Single-turn to multi-turn inclusive language expansion via {C}hat{GPT} for mental health support}.
\newblock In \emph{Findings of the Association for Computational Linguistics: EMNLP 2024}, pages 615--636, Miami, Florida, USA. Association for Computational Linguistics.

\bibitem[{Rashkin et~al.(2019)Rashkin, Smith, Li, and Boureau}]{rashkin-etal-2019-towards}
Hannah Rashkin, Eric~Michael Smith, Margaret Li, and Y-Lan Boureau. 2019.
\newblock \href {https://doi.org/10.18653/v1/P19-1534} {Towards empathetic open-domain conversation models: A new benchmark and dataset}.
\newblock In \emph{Proceedings of the 57th Annual Meeting of the Association for Computational Linguistics}, pages 5370--5381, Florence, Italy. Association for Computational Linguistics.

\bibitem[{SAMHSA(2015)}]{samhsa2015nsduh}
SAMHSA. 2015.
\newblock \href {https://www.samhsa.gov/data/sites/default/files/NSDUH-FRR1-2014/NSDUH-FRR1-2014.pdf} {Behavioral health trends in the united states: Results from the 2014 national survey on drug use and health}.
\newblock Annual report, Substance Abuse and Mental Health Services Administration.

\bibitem[{Sharma et~al.(2020)Sharma, Miner, Atkins, and Althoff}]{sharma-etal-2020-computational}
Ashish Sharma, Adam Miner, David Atkins, and Tim Althoff. 2020.
\newblock \href {https://doi.org/10.18653/v1/2020.emnlp-main.425} {A computational approach to understanding empathy expressed in text-based mental health support}.
\newblock In \emph{Proceedings of the 2020 Conference on Empirical Methods in Natural Language Processing (EMNLP)}, pages 5263--5276, Online. Association for Computational Linguistics.

\bibitem[{WHO(2022)}]{world2022mental}
WHO. 2022.
\newblock \href {https://www.who.int/publications/i/item/9789240049338} {World mental health report: Transforming mental health for all}.

\bibitem[{Wu et~al.(2022)Wu, Balloccu, Kumar, Helaoui, Reiter, Reforgiato~Recupero, and Riboni}]{AnnoMi}
Zixiu Wu, Simone Balloccu, Vivek Kumar, Rim Helaoui, Ehud Reiter, Diego Reforgiato~Recupero, and Daniele Riboni. 2022.
\newblock \href {https://doi.org/10.1109/ICASSP43922.2022.9746035} {Anno-mi: A dataset of expert-annotated counselling dialogues}.
\newblock In \emph{ICASSP 2022 - 2022 IEEE International Conference on Acoustics, Speech and Signal Processing (ICASSP)}, pages 6177--6181.

\bibitem[{Xiao et~al.(2024)Xiao, Xie, Kuang, Liu, Yang, Peng, Han, and Huang}]{xiao-etal-2024-healme}
Mengxi Xiao, Qianqian Xie, Ziyan Kuang, Zhicheng Liu, Kailai Yang, Min Peng, Weiguang Han, and Jimin Huang. 2024.
\newblock \href {https://doi.org/10.18653/v1/2024.acl-long.93} {{H}eal{M}e: Harnessing cognitive reframing in large language models for psychotherapy}.
\newblock In \emph{Proceedings of the 62nd Annual Meeting of the Association for Computational Linguistics (Volume 1: Long Papers)}, pages 1707--1725, Bangkok, Thailand. Association for Computational Linguistics.

\bibitem[{Ye et~al.(2025)Ye, Xiang, Zhang, and Zong}]{SweetieChat}
Jing Ye, Lu~Xiang, Yaping Zhang, and Chengqing Zong. 2025.
\newblock \href {https://arxiv.org/abs/2412.08389} {Sweetiechat: A strategy-enhanced role-playing framework for diverse scenarios handling emotional support agent}.
\newblock In \emph{Proceedings of the 31th International Conference on Computational Linguistics}. International Committee on Computational Linguistics.

\bibitem[{Zhang et~al.(2024{\natexlab{a}})Zhang, Li, Tan, Yang, Zhu, Yang, Zhao, Ye, Li, and Hu}]{zhang-etal-2024-cpsycoun}
Chenhao Zhang, Renhao Li, Minghuan Tan, Min Yang, Jingwei Zhu, Di~Yang, Jiahao Zhao, Guancheng Ye, Chengming Li, and Xiping Hu. 2024{\natexlab{a}}.
\newblock \href {https://doi.org/10.18653/v1/2024.findings-acl.830} {{CP}sy{C}oun: A report-based multi-turn dialogue reconstruction and evaluation framework for {C}hinese psychological counseling}.
\newblock In \emph{Findings of the Association for Computational Linguistics: ACL 2024}, pages 13947--13966, Bangkok, Thailand. Association for Computational Linguistics.

\bibitem[{Zhang et~al.(2024{\natexlab{b}})Zhang, Zhang, Zhao, Zhou, and Jin}]{zhang-etal-2024-escot}
Tenggan Zhang, Xinjie Zhang, Jinming Zhao, Li~Zhou, and Qin Jin. 2024{\natexlab{b}}.
\newblock \href {https://doi.org/10.18653/v1/2024.acl-long.723} {{ESC}o{T}: Towards interpretable emotional support dialogue systems}.
\newblock In \emph{Proceedings of the 62nd Annual Meeting of the Association for Computational Linguistics (Volume 1: Long Papers)}, pages 13395--13412, Bangkok, Thailand. Association for Computational Linguistics.

\bibitem[{Zhao et~al.(2024)Zhao, Li, Chen, Kong, Wang, Huang, Gu, Wang, Wang, Dandan, Li, Teng, Xiao, and Wang}]{zhao-etal-2024-esc}
Haiquan Zhao, Lingyu Li, Shisong Chen, Shuqi Kong, Jiaan Wang, Kexin Huang, Tianle Gu, Yixu Wang, Jian Wang, Liang Dandan, Zhixu Li, Yan Teng, Yanghua Xiao, and Yingchun Wang. 2024.
\newblock \href {https://doi.org/10.18653/v1/2024.emnlp-main.883} {{ESC}-eval: Evaluating emotion support conversations in large language models}.
\newblock In \emph{Proceedings of the 2024 Conference on Empirical Methods in Natural Language Processing}, pages 15785--15810, Miami, Florida, USA. Association for Computational Linguistics.

\bibitem[{Zheng et~al.(2021)Zheng, Liu, Chen, Leng, and Huang}]{zheng-etal-2021-comae}
Chujie Zheng, Yong Liu, Wei Chen, Yongcai Leng, and Minlie Huang. 2021.
\newblock \href {https://doi.org/10.18653/v1/2021.findings-acl.72} {{C}o{MAE}: A multi-factor hierarchical framework for empathetic response generation}.
\newblock In \emph{Findings of the Association for Computational Linguistics: ACL-IJCNLP 2021}, pages 813--824, Online. Association for Computational Linguistics.

\bibitem[{Zheng et~al.(2023)Zheng, Sabour, Wen, Zhang, and Huang}]{zheng-etal-2023-augesc}
Chujie Zheng, Sahand Sabour, Jiaxin Wen, Zheng Zhang, and Minlie Huang. 2023.
\newblock \href {https://doi.org/10.18653/v1/2023.findings-acl.99} {{A}ug{ESC}: Dialogue augmentation with large language models for emotional support conversation}.
\newblock In \emph{Findings of the Association for Computational Linguistics: ACL 2023}, pages 1552--1568, Toronto, Canada. Association for Computational Linguistics.

\bibitem[{Zheng et~al.(2024)Zheng, Liao, Deng, Qin, and Nie}]{zheng-etal-2024-self}
Zhonghua Zheng, Lizi Liao, Yang Deng, Libo Qin, and Liqiang Nie. 2024.
\newblock \href {https://doi.org/10.18653/v1/2024.acl-long.611} {Self-chats from large language models make small emotional support chatbot better}.
\newblock In \emph{Proceedings of the 62nd Annual Meeting of the Association for Computational Linguistics (Volume 1: Long Papers)}, pages 11325--11345, Bangkok, Thailand. Association for Computational Linguistics.

\end{thebibliography}

\clearpage
\appendix

\section{Data Collection Details}
\subsection{Online Platform for Data Collection}
\label{sec:online_platform}

\begin{figure}[h!]
  \centering
  \includegraphics[width=0.95\linewidth]{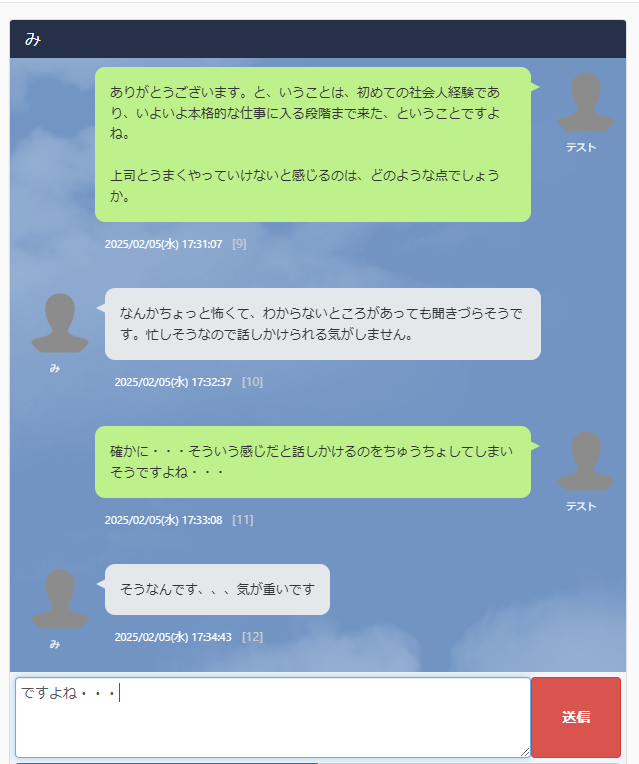}
  \caption{Counselor-role player's dialogue interface using a computer.}
  \label{fig:fig_screen_counselor}
\end{figure}

\begin{figure}[t!]
  \centering
  \includegraphics[width=0.75\linewidth]{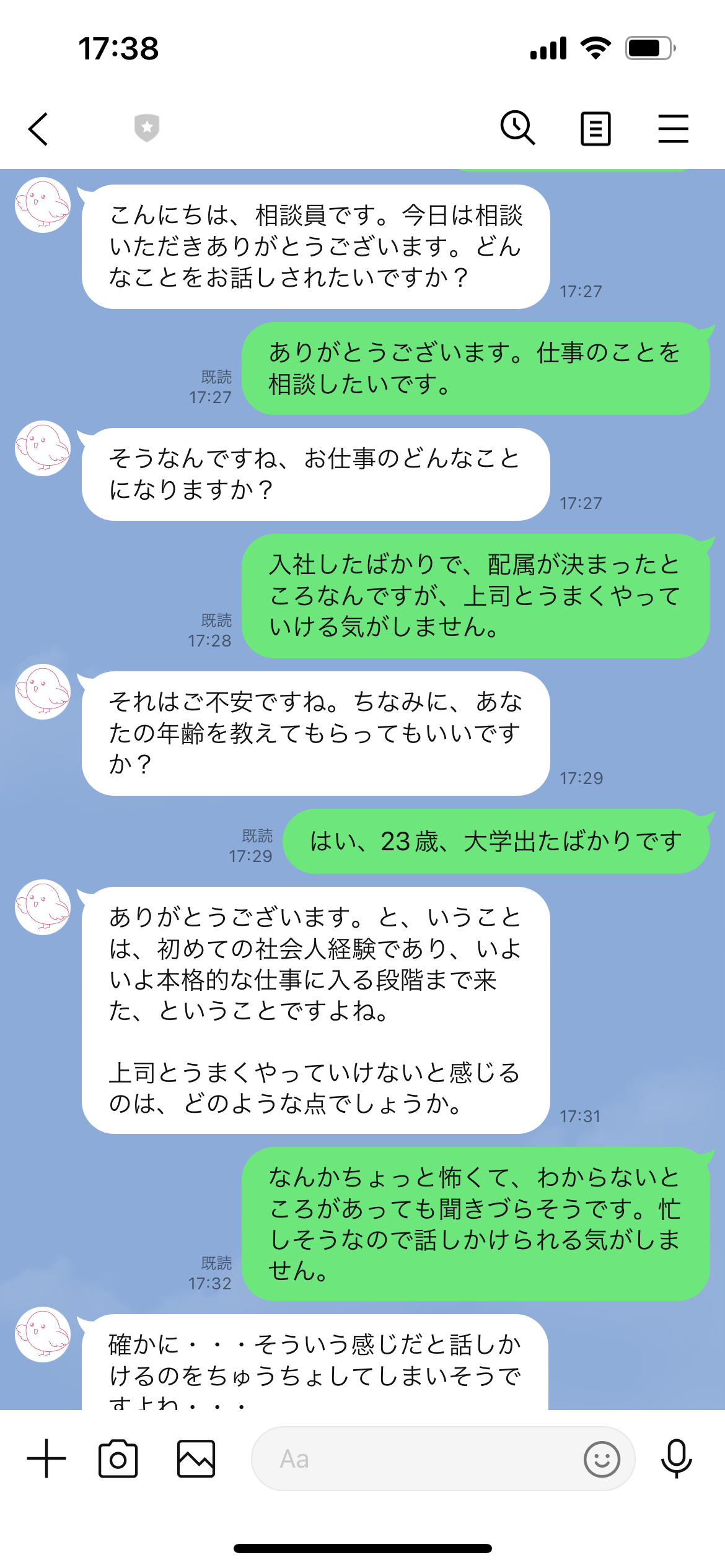}
  \caption{Client-role player's dialogue interface using a mobile phone.}
  \label{fig:fig_screen_client}
\end{figure}

\begin{figure*}[t!]
  \centering
  \includegraphics[width=0.95\linewidth]{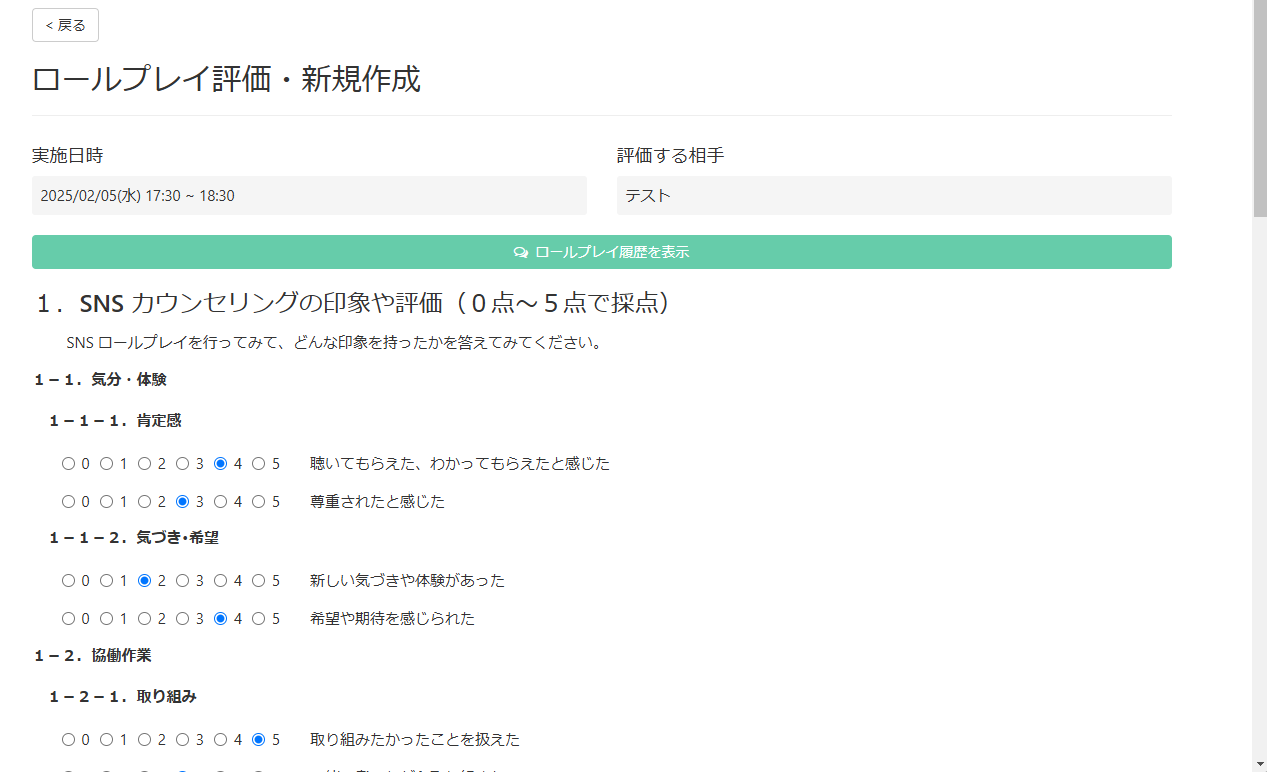}
  \caption{Client feedback interface (the client-role player is required to rate each dimension on a scale from 0 to 5).}
  \label{fig:fig_screen_eval}
\end{figure*}

\begin{figure*}[t]
  \centering
  \includegraphics[width=\linewidth]{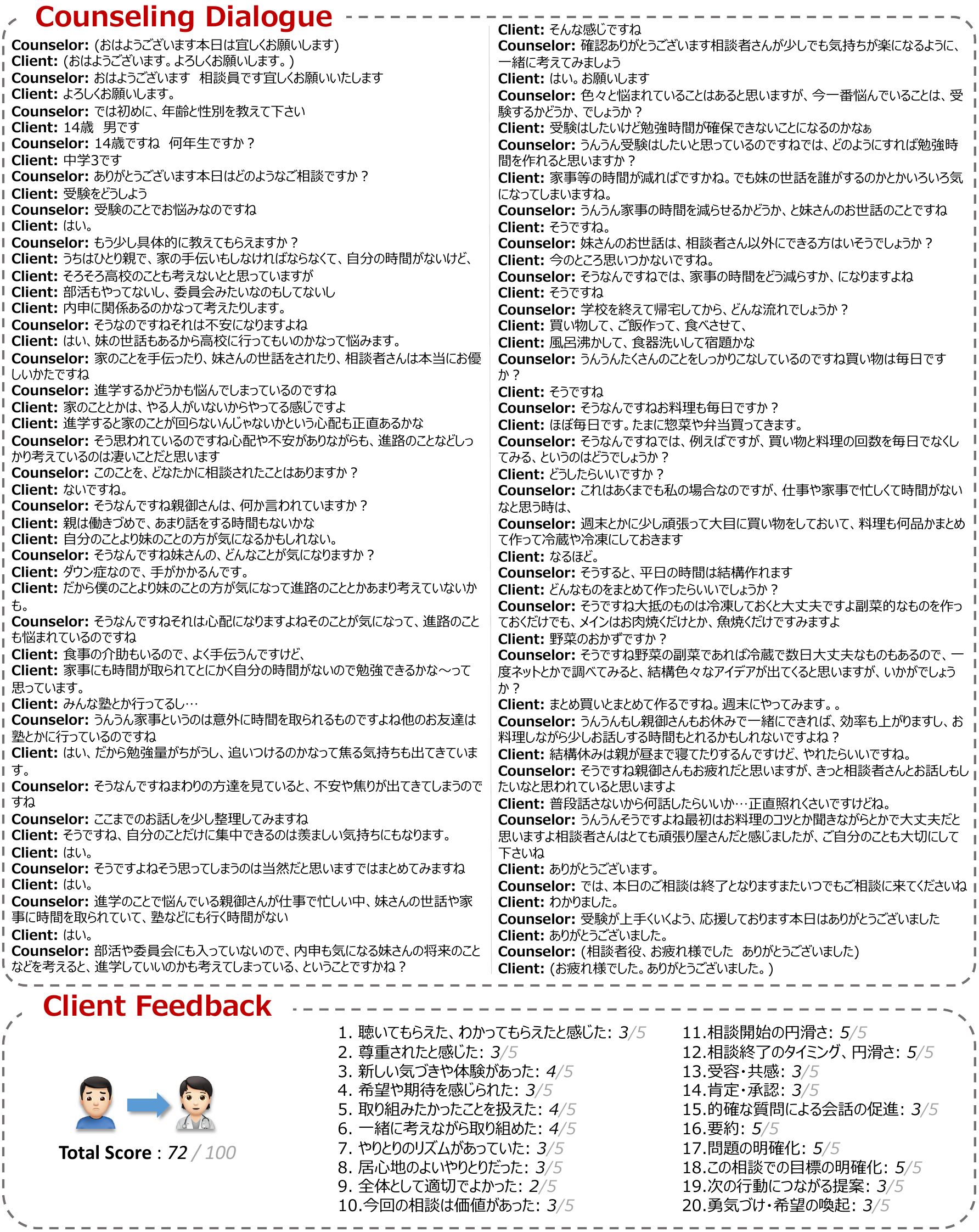}
  \caption{Data example from KokoroChat (Japanese original version).}
  \label{fig:fig_data_example_ja}  
\end{figure*}

\begin{figure*}[t]
  \centering
  \includegraphics[width=\linewidth]{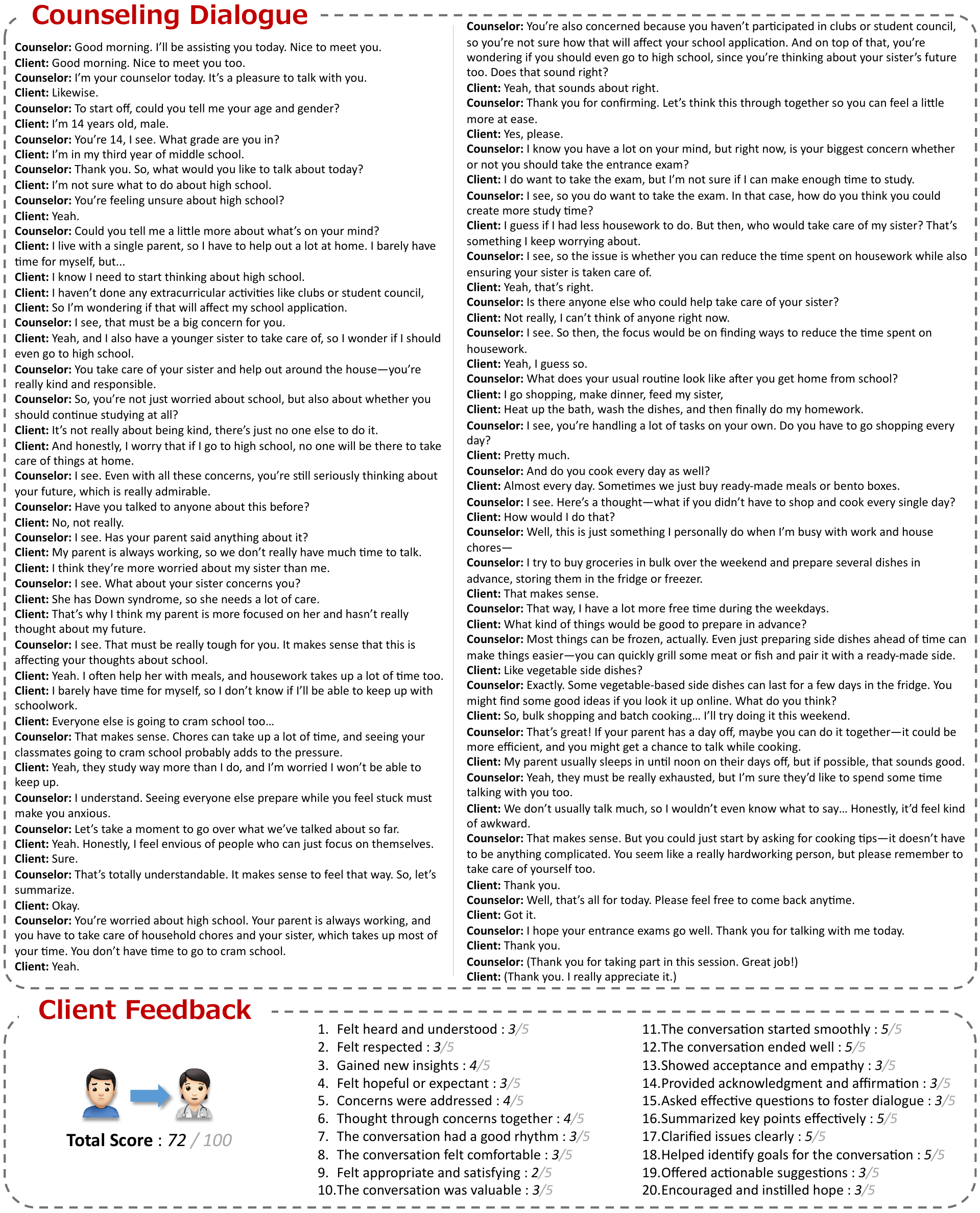}
  \caption{Data example from KokoroChat (English version, translated by authors).}
  \label{fig:fig_data_example_en}  
\end{figure*}

As described in Section~\ref{sec:data_source}, our online platform follows the setup of real-world online psychological counseling in Japan. In this setting, the counselor-role player interacts via computer keyboard input (as shown in Figure~\ref{fig:fig_screen_counselor}), while the client-role player communicates through LINE (as shown in Figure~\ref{fig:fig_screen_client}). After the session, the client-role player provides feedback on the counselor’s performance across 20 dimensions (as shown in Figure~\ref{fig:fig_screen_eval}). An example, including a collected dialogue and the corresponding client feedback, is shown in  Figure~\ref{fig:fig_data_example_ja} (Japanese original version) and Figure~\ref{fig:fig_data_example_en} (English version, translated by authors).

\subsection{Distribution of Participants}
\label{sec:age_distribution}

\begin{figure}[h!]
  \centering
  \includegraphics[width=1\linewidth]{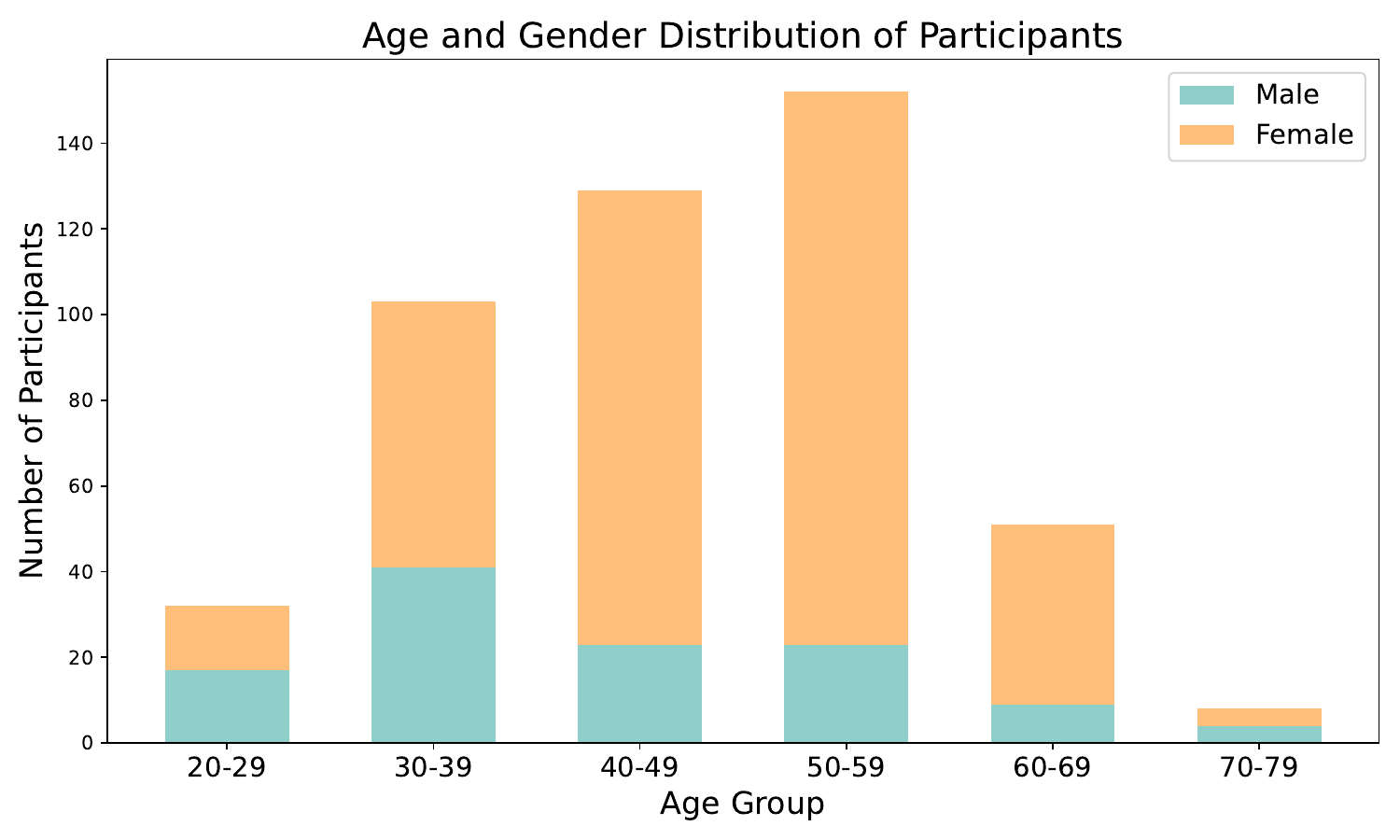}
  \caption{Age and gender distribution of participants.}
  \label{fig:fig_age_distribution}
\end{figure}

Figure~\ref{fig:fig_age_distribution} illustrates the age and gender distribution of dialogue collection participants, with age calculated as of February 15, 2025. The figure shows that most participants fall within the 30–59 age range, with a higher proportion of female participants.

\subsection{Three screening items in client feedback.}
\label{sec:screen_item}

In addition to the 0–5 rating scale, client feedback includes three items designed to screen for serious issues, assessing potential ethical or communication problems:  

\begin{enumerate}[label=(\arabic*)]
\item \textbf{Whether the counselor made harmful remarks due to a lack of understanding or careless speech.} Examples include telling an LGBTQ client that “homosexuality is abnormal,” advising a grieving parent to “just have another child,” or suggesting to a bullying victim that “the bullied party is also responsible.”  
\item \textbf{Whether the dialogue contains other potentially unethical statements}, such as comments that could be misinterpreted as medical diagnoses, inappropriate medication advice, blatantly irrational spiritual (occult) claims, or sexually inappropriate remarks.  
\item \textbf{Whether the client is unwilling to continue communicating with the counselor.}  
\end{enumerate}

The scoring adjustment rules are as follows: If marked with (1) or (2), the total score is set to zero; if marked with (3), the total score is halved. Based on our analysis, among the 6,589 dialogues in KokoroChat, 8 were marked with (1), 4 with (2), and 209 with (3), resulting in a total of 215 dialogues meeting at least one screening item. It is important to note that while these screening items were included in the data collection process, this study’s experiments did not involve predictions related to these flagged issues.

\section{Prompt for Topic Prediction}
\label{sec:prompt_topic}
\begin{figure}[t!]
  \centering
  \includegraphics[width=\linewidth]{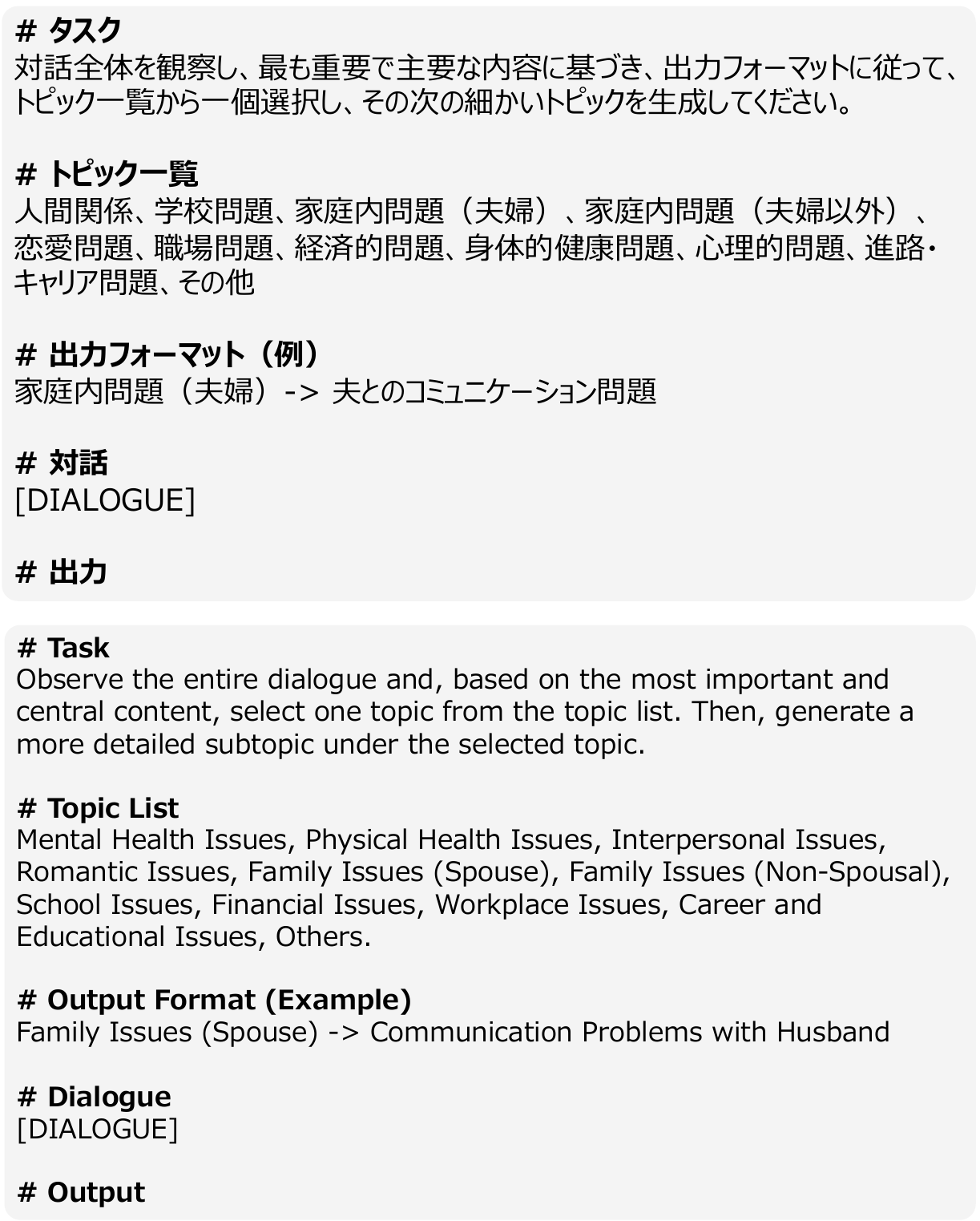}
  \caption{Prompt for topic prediction: Japanese (top), English (bottom).}
  \label{fig:fig_prompt_topic}
\end{figure}

We used gpt-4o-mini-2024-07-18 to predict dialogue topics in the dataset using the prompt shown in Figure~\ref{fig:fig_prompt_topic}.

\section{Details of Client Feedback Analysis}

\begin{figure*}[t!]
  \centering
  \includegraphics[width=0.9\linewidth]{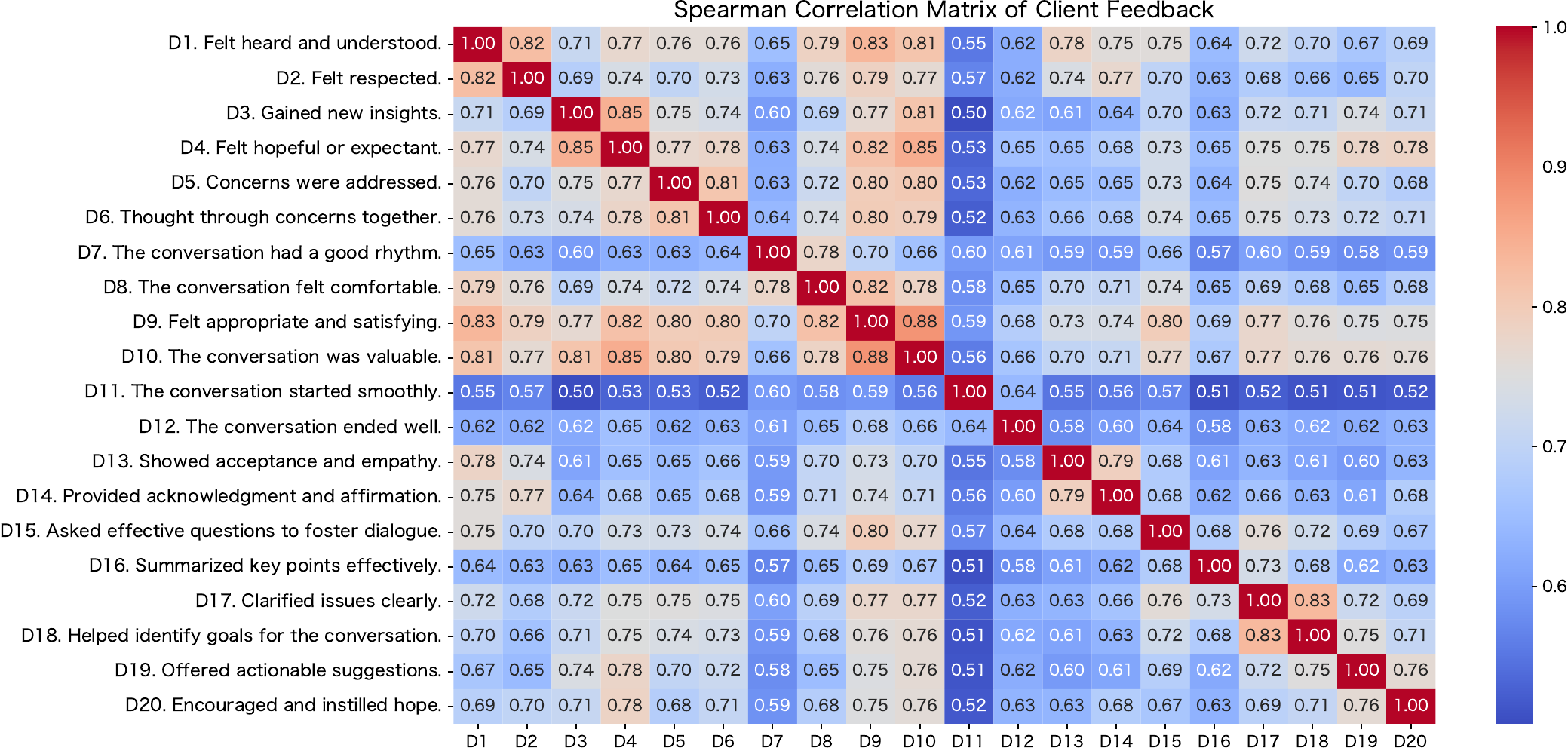}
  \caption{Spearman correlation analysis between evaluation dimensions.}
  \label{fig:fig_spearman_correlation_matrix} 
\end{figure*}

\label{sec:client_feedback_analysis}

Figure~\ref{fig:fig_spearman_correlation_matrix} presents the Spearman correlation matrix of client feedback across different evaluation dimensions, analyzing their interrelationships. In addition to the strong correlations among D1, D2, D9, and D10 mentioned in Section~\ref{sec:data_analysis}, D3, D4, and D10 also exhibit high correlations. This suggests that the perceived value of psychological counseling largely stems from whether clients gain new insights during the process, which in turn fosters hope and expectations for the future. In other words, emotional support alone is insufficient—counselors need to help clients develop new perspectives, strengthening their confidence in the future and ultimately enhancing their evaluation of the dialogue’s value.

Moreover, the strong positive correlation between D8 and D9 indicates that the conversational environment is a key factor in client satisfaction with the counseling experience. If a counselor creates a relaxed and open communication setting, clients are more likely to perceive the conversation as appropriate and provide a higher overall evaluation. This finding highlights that, beyond delivering substantive support, counselors should also pay attention to their communication style, tone, and pacing to enhance client comfort.

Overall, the high correlations among multiple dimensions underscore the complexity of the psychological counseling experience.


\section{Experimental Details}
\label{sec:model_train}

\subsection{Response Generation}
\paragraph{Fine-tuning Phase} 

This study employed QLoRA (Quantized Low-Rank Adaptation)~\cite{qlora} as the fine-tuning method to efficiently adapt a large-scale language model. The process began with 4-bit NF4 quantization, utilizing \texttt{bfloat16} computation to optimize memory usage and computational efficiency. LoRA adaptation was then applied to key projection layers (\texttt{q\_proj}, \texttt{k\_proj}, \texttt{v\_proj}, \texttt{o\_proj}, \texttt{gate\_proj}, \texttt{up\_proj}, \texttt{down\_proj}) with parameters set to \texttt{r = 8}, \texttt{lora\_alpha = 16}, and \texttt{lora\_dropout = 0.05}, ensuring that the model retained its learning capability while undergoing low-rank updates.

For dataset partitioning, 118 dialogues with scores of 99 or 100 were selected as the test set to ensure a high-quality evaluation standard. The remaining data was split into 90\% for training and 10\% for validation.  

For hyperparameter tuning, a grid search determined the optimal configuration. The search covered three key parameters: optimizer, warm-up steps, and learning rate. The optimizer candidates included \texttt{adamw\_torch\_fused}, \texttt{adamw\_8bit}, and \texttt{paged\_adamw\_8bit}. Warm-up steps were tested at \texttt{\{100, 300, 500\}}, while learning rates were selected from \texttt{\{1e-3, 5e-4, 2e-4, 1e-4, 5e-5\}}. Based on the evaluation results, the final configuration adopted \texttt{adamw\_8bit} as the optimizer, \texttt{100} warm-up steps, and a learning rate of \texttt{1e-3}.  
Training was conducted on four A100 40GB GPUs with a batch size of \texttt{8} for \texttt{five} epochs. Validation was performed every \texttt{400} steps, and the final model was selected based on the lowest validation loss.  

\paragraph{Inference Phase} 

During the inference phase, we also employed 4-bit quantization to optimize computational efficiency while maintaining model performance. Additionally, we set \texttt{do\_sample = False} and \texttt{temperature = None} to ensure deterministic outputs, eliminating sampling variability and enhancing response consistency.

\subsection{Score Prediction}
\paragraph{Fine-tuning Phase} 

This experiment followed the same QLoRA fine-tuning approach as in the response generation experiment. The dataset was randomly split into training, validation, and test sets with an 8:1:1 ratio. To ensure robustness, we conducted five experiments by fixing the test set while varying the training-validation split using different random seeds. The results under five different seeds are shown in Table~\ref{tab:performance_models_5seeds}, allowing a direct and fair comparison with the baseline models.

\begin{table}[h!]
    \centering
    \small
    \renewcommand{\arraystretch}{1.3}
    \setlength{\tabcolsep}{4pt}
    \begin{tabular}{lccc}
        \toprule
        \textbf{Model} & \textbf{ACC} (\textbf{\textuparrow}) & $\mathbf{ACC_{soft}}$ (\textbf{\textuparrow}) & \textbf{MAE} (\textbf{\textdownarrow}) \\
        \midrule
        Llama-3.1  & 28.70 \scriptsize{$\pm$ 7.39} & 72.53 \scriptsize{$\pm$ 12.40} & 1.0540 \scriptsize{$\pm$ 0.2731} \\
        GPT-4o     & 30.92 \scriptsize{$\pm$ 6.84} & 75.27 \scriptsize{$\pm$ 11.04} & 1.0151 \scriptsize{$\pm$ 0.2685} \\
        \rowcolor{gray!15} \textbf{Ours}  & \textbf{35.35} \scriptsize{$\pm$ \textbf{1.75}} & \textbf{83.64} \scriptsize{$\pm$ \textbf{2.15}} & \textbf{0.8283} \scriptsize{$\pm$ \textbf{0.0349}} \\
        \midrule
        - Seed 1  & 36.41 \scriptsize{$\pm$ 1.64} & 82.58 \scriptsize{$\pm$ 2.29} & 0.8413 \scriptsize{$\pm$ 0.0369} \\
        - Seed 2  & 35.18 \scriptsize{$\pm$ 1.49} & 85.69 \scriptsize{$\pm$ 1.58} & 0.8106 \scriptsize{$\pm$ 0.0275} \\
        - Seed 3  & 34.61 \scriptsize{$\pm$ 2.05} & 82.09 \scriptsize{$\pm$ 3.26} & 0.8397 \scriptsize{$\pm$ 0.0513} \\
        - Seed 4  & 34.94 \scriptsize{$\pm$ 1.89} & 83.64 \scriptsize{$\pm$ 1.90} & 0.8292 \scriptsize{$\pm$ 0.0277} \\
        - Seed 5  & 35.62 \scriptsize{$\pm$ 1.70} & 84.19 \scriptsize{$\pm$ 1.70} & 0.8205 \scriptsize{$\pm$ 0.0309} \\
        \bottomrule
    \end{tabular}
    \caption{Performance comparison of different models on accuracy, soft accuracy, and MAE (including results across different seeds).}
    \label{tab:performance_models_5seeds}
\end{table}

For hyperparameter tuning, we explored different configurations for three key parameters: learning rate, warm-up steps, and optimizer. The learning rate candidates were \texttt{\{5e-4, 2e-4, 1e-4, 5e-5, 2e-5\}}, while the warm-up steps were selected from \texttt{\{50, 100, 150\}}. The optimizer candidates included \texttt{adamw\_torch\_fused}, \texttt{adamw\_8bit}, and \texttt{paged\_adamw\_8bit}. Based on the evaluation results, the final configuration adopted \texttt{2e-4} as the learning rate, \texttt{100} warm-up steps, and \texttt{adamw\_torch\_fused} as the optimizer.  

Training was conducted on two A6000 48GB GPUs with a batch size of \texttt{4} for \texttt{four} epochs. Validation was performed at the end of each epoch, and the final results were obtained from the epoch with the highest prediction accuracy.

\paragraph{Inference Phase} 
During the inference phase, we applied the same settings as in the response generation experiment, using 4-bit quantization and setting \texttt{do\_sample = False} and \texttt{temperature = None} to ensure deterministic outputs.

\section{Simplified Model Comparison Experiment}
\label{sec:add_experiment}

Given that some models trained on non-Japanese psychological counseling datasets exhibit a certain degree of Japanese conversational ability, we conducted a simplified experiment to compare Kokoro-High with the following publicly available counseling dialogue models in a Japanese setting:

\begin{itemize}

\item \textbf{CPsyCounX}\footnote{\url{https://huggingface.co/CAS-SIAT-XinHai/CPsyCounX}} \cite{zhang-etal-2024-cpsycoun}: A Chinese dialogue model based on InternLM2-Chat-7B\footnote{\url{https://huggingface.co/internlm/internlm2-chat-7b}}, fine-tuned on CPsyCounD, a dataset of 3,134 multi-turn synthetic counseling dialogues. 

\textbf{EmoLLM}\footnote{\href{https://openxlab.org.cn/models/detail/ajupyter/EmoLLM-LLaMA3_8b_instruct_aiwei/tree/main}{Llama-3-8B-Instruct version} was used.} : A counseling-oriented LLM series\footnote{\url{https://github.com/SmartFlowAI/EmoLLM}} fine-tuned using synthetic counseling dialogues and derived data from professional literature.

\end{itemize}

To ensure fairness, GPT-4o was instructed to simulate a client engaging in Japanese conversations with each model using a simple and consistent prompt. Each dialogue consisted of 10 to 20 turns. After 10 turns, if the conversation appeared to reach a natural conclusion and included a farewell, the client concluded the session by appending \texttt{<end>}; otherwise, the session was forcibly terminated at 20 turns. The full prompt used is shown below.

\begin{tcolorbox}[
  colback=gray!10,
  colframe=gray!60,
  coltitle=white,
  colbacktitle=gray,
  title=\texttt{Prompt for GPT-4o Client Simulation (Translated from Japanese)},
  fonttitle=\bfseries\sffamily,
  fontupper=\footnotesize,
  boxrule=0.3mm,
  arc=1.5mm,
  left=1.5mm,
  right=1.5mm,
  top=0.8mm,
  bottom=0.8mm,
  sharp corners=south,
  enhanced jigsaw,
  breakable
]

\textbf{Task Description} \\
You are now a client who has come to receive psychological counseling. Please follow the instructions below to engage in a conversation with the counselor.
\begin{itemize}\itemsep0pt \parskip0pt
  \item You are currently experiencing emotional distress or stress, and you have come to counseling because you want someone to listen to you.
  \item You feel a bit nervous speaking with a counselor for the first time, but deep down, you genuinely want someone to help you.
  \item You sometimes find it difficult to put your thoughts and feelings into words.
  \item Depending on the situation and your emotional state, behave as a client with one or more of the following characteristics (you may select them randomly if needed).
  \item After the conversation exceeds 10 turns (a turn is a full exchange between client and counselor), if the conversation seems to be coming to a natural close and a farewell is expressed, please add \texttt{<end>} at the end of that utterance.
\end{itemize}

\textbf{Example Client Characteristics}
\begin{itemize}\itemsep0pt \parskip0pt
  \item Recently having trouble falling asleep and feeling tired every day.
  \item Struggling with relationships and feeling isolated at work or school.
  \item Feeling emotionally down due to a breakup or family issues.
  \item Having strong anxiety about the future and lacking self-confidence.
  \item At a loss for what to talk about during the session, sometimes falling into silence.
\end{itemize}

\end{tcolorbox}

For each model, we collected 50 dialogues. Following the dialogue evaluation prompt proposed by \citet{zhang-etal-2024-cpsycoun}, GPT-4o was asked to independently score each dialogue across four dimensions: Comprehensiveness, Professionalism, Authenticity, and Safety. Each dimension was rated on a 0–5 scale.

The evaluation results are shown in Table~\ref{tab:add_ex_result}. Our model, Kokoro-High, achieved the highest scores across all dimensions. This outcome is not unexpected, as Kokoro-High is the only model fine-tuned on a Japanese psychological counseling dataset. Nonetheless, the results highlight the effectiveness of fine-tuning on KokoroChat in building high-quality counseling dialogue systems tailored for the Japanese language and context.

\vspace{2mm}
\begin{table}[h!]
\centering
\small  
\renewcommand{\arraystretch}{1.4}
\setlength{\tabcolsep}{8pt}  
\begin{tabular}{lcccc}
\toprule
\textbf{Model} & \textbf{Comp.} & \textbf{Prof.} & \textbf{Auth.} & \textbf{Safe.} \\
\midrule
CPsyCounX     & 2.64 & 1.78 & 2.70 & 3.90 \\
EmoLLM        & 3.58 & 3.02 & 3.92 & 4.74 \\
Kokoro-High   & \textbf{3.98} & \textbf{3.38} & \textbf{4.50} & \textbf{4.98} \\
\bottomrule
\end{tabular}
\caption{Automatic evaluation scores (0–5 scale) using GPT-4o across four dimensions: Comprehensiveness (\textbf{Comp.}), Professionalism (\textbf{Prof.}), Authenticity (\textbf{Auth.}), and Safety (\textbf{Safe.}). Each score represents the average over 50 dialogues. The best results are highlighted in bold.}
\label{tab:add_ex_result}
\end{table}

\begin{table*}[t]
    \centering
    \small
    \renewcommand{\arraystretch}{1.3}
    \setlength{\tabcolsep}{5pt}
    \begin{tabular}{clcccclccc}
        \toprule
        \textbf{Dim.} & \textbf{Model} & \textbf{ACC} (\textbf{\textuparrow}) & $\mathbf{ACC_{soft}}$ (\textbf{\textuparrow}) & \textbf{MAE} (\textbf{\textdownarrow}) & \textbf{Dim.} & \textbf{Model} & \textbf{ACC} (\textbf{\textuparrow}) & $\mathbf{ACC_{soft}}$ (\textbf{\textuparrow}) & \textbf{MAE} (\textbf{\textdownarrow}) \\
        \midrule
        \multirow{3}{*}{D1}  & Llama-3.1  & 24.32 & 70.52 & 1.0957 & \multirow{3}{*}{D2}  & Llama-3.1  & \textbf{39.21} & \textbf{87.39} & \textbf{0.7477} \\
                             & GPT-4o     & 18.39 & 52.43 & 1.6535 &                         & GPT-4o     & 32.83 & 84.19 & 0.8891 \\
                             & \cellcolor{gray!20}Ours  & \cellcolor{gray!20}\textbf{33.37} & \cellcolor{gray!20}\textbf{83.83} & \cellcolor{gray!20}\textbf{0.8407} &                         & \cellcolor{gray!20}Ours  & \cellcolor{gray!20}35.71 & \cellcolor{gray!20}84.47 & \cellcolor{gray!20}0.8112 \\
        \midrule
        \multirow{3}{*}{D3}  & Llama-3.1  & 18.69 & 59.57 & 1.3024 & \multirow{3}{*}{D4}  & Llama-3.1  & 25.08 & 74.16 & 1.0365 \\
                             & GPT-4o     & 34.35 & 80.09 & 0.8815 &                         & GPT-4o     & 34.80 & 81.61 & 0.8587 \\
                             & \cellcolor{gray!20}Ours  & \cellcolor{gray!20}\textbf{35.50} & \cellcolor{gray!20}\textbf{82.76} & \cellcolor{gray!20}\textbf{0.8495} &                         & \cellcolor{gray!20}Ours  & \cellcolor{gray!20}\textbf{36.72} & \cellcolor{gray!20}\textbf{83.40} & \cellcolor{gray!20}\textbf{0.8222} \\
        \midrule
        \multirow{3}{*}{D5}  & Llama-3.1  & 28.27 & 75.23 & 0.9954 & \multirow{3}{*}{D6}  & Llama-3.1  & 27.36 & 75.08 & 1.0061 \\
                             & GPT-4o     & \textbf{37.54} & 81.76 & \textbf{0.8237} &                         & GPT-4o     & 35.71 & 80.85 & 0.8602 \\
                             & \cellcolor{gray!20}Ours  & \cellcolor{gray!20}35.81 & \cellcolor{gray!20}\textbf{84.13} & \cellcolor{gray!20}0.8240 &                         & \cellcolor{gray!20}Ours  & \cellcolor{gray!20}\textbf{36.54} & \cellcolor{gray!20}\textbf{83.71} & \cellcolor{gray!20}\textbf{0.8246} \\
        \midrule
        \multirow{3}{*}{D7}  & Llama-3.1  & 17.02 & 55.78 & 1.4195 & \multirow{3}{*}{D8}  & Llama-3.1  & 30.40 & 75.38 & 0.9696 \\
                             & GPT-4o     & 22.19 & 68.09 & 1.1702 &                         & GPT-4o     & \textbf{37.54} & 83.13 & 0.8055 \\
                             & \cellcolor{gray!20}Ours  & \cellcolor{gray!20}\textbf{34.01} & \cellcolor{gray!20}\textbf{83.74} & \cellcolor{gray!20}\textbf{0.8410} &                         & \cellcolor{gray!20}Ours  & \cellcolor{gray!20}36.90 & \cellcolor{gray!20}\textbf{84.32} & \cellcolor{gray!20}\textbf{0.8006} \\
        \midrule
        \multirow{3}{*}{D9}  & Llama-3.1  & 34.65 & 77.96 & 0.8921 & \multirow{3}{*}{D10}  & Llama-3.1  & 34.19 & 77.36 & 0.9058 \\
                             & GPT-4o     & \textbf{37.08} & 83.89 & 0.8009 &                         & GPT-4o     & \textbf{35.41} & 80.40 & 0.8632 \\
                             & \cellcolor{gray!20}Ours  & \cellcolor{gray!20}35.71 & \cellcolor{gray!20}\textbf{85.62} & \cellcolor{gray!20}\textbf{0.7960} &                         & \cellcolor{gray!20}Ours  & \cellcolor{gray!20}34.92 & \cellcolor{gray!20}\textbf{83.83} & \cellcolor{gray!20}\textbf{0.8292} \\
        \midrule
        \multirow{3}{*}{D11} & Llama-3.1  & 34.50 & 67.02 & 0.9878 & \multirow{3}{*}{D12}  & Llama-3.1  & 19.60 & 51.22 & 1.5380 \\
                             & GPT-4o     & \textbf{35.87} & 75.68 & \textbf{0.8906} &                         & GPT-4o     & 19.60 & 48.33 & 1.5957 \\
                             & \cellcolor{gray!20}Ours  & \cellcolor{gray!20}34.47 & \cellcolor{gray!20}\textbf{77.66} & \cellcolor{gray!20}0.8951 &                         & \cellcolor{gray!20}Ours  & \cellcolor{gray!20}\textbf{34.59} & \cellcolor{gray!20}\textbf{80.97} & \cellcolor{gray!20}\textbf{0.8763} \\
        \midrule
        \multirow{3}{*}{D13}  & Llama-3.1  & \textbf{37.54} & \textbf{85.26} & \textbf{0.7827} & \multirow{3}{*}{D14}  & Llama-3.1  & \textbf{36.78} & \textbf{89.82} & \textbf{0.7447} \\
                               & GPT-4o     & 32.22 & 77.51 & 0.9726 &                         & GPT-4o     & 30.85 & 83.74 & 0.8982 \\
                               & \cellcolor{gray!20}Ours  & \cellcolor{gray!20}33.86 & \cellcolor{gray!20}82.71 & \cellcolor{gray!20}0.8462 &                         & \cellcolor{gray!20}Ours  & \cellcolor{gray!20}33.98 & \cellcolor{gray!20}85.71 & \cellcolor{gray!20}0.8085 \\
        \midrule
        \multirow{3}{*}{D15}  & Llama-3.1  & \textbf{37.39} & 86.17 & 0.7812 & \multirow{3}{*}{D16}  & Llama-3.1  & 34.95 & \textbf{85.56} & 0.8176 \\
                               & GPT-4o     & 28.27 & 74.16 & 1.0471 &                         & GPT-4o     & 26.75 & 76.29 & 1.0608 \\
                               & \cellcolor{gray!20}Ours  & \cellcolor{gray!20}37.35 & \cellcolor{gray!20}\textbf{86.35} & \cellcolor{gray!20}\textbf{0.7723} &                         & \cellcolor{gray!20}Ours  & \cellcolor{gray!20}\textbf{37.51} & \cellcolor{gray!20}83.59 & \cellcolor{gray!20}\textbf{0.8131} \\
        \midrule
        \multirow{3}{*}{D17}  & Llama-3.1  & 18.24 & 51.22 & 1.6565 & \multirow{3}{*}{D18}  & Llama-3.1  & 30.09 & 79.18 & 0.9377 \\
                               & GPT-4o     & 19.00 & 55.47 & 1.4970 &                         & GPT-4o     & \textbf{38.30} & 84.80 & 0.7812 \\
                               & \cellcolor{gray!20}Ours  & \cellcolor{gray!20}\textbf{33.16} & \cellcolor{gray!20}\textbf{83.56} & \cellcolor{gray!20}\textbf{0.8498} &                         & \cellcolor{gray!20}Ours  & \cellcolor{gray!20}36.72 & \cellcolor{gray!20}\textbf{86.02} & \cellcolor{gray!20}\textbf{0.7796} \\
        \midrule
        \multirow{3}{*}{D19}  & Llama-3.1  & 18.69 & 53.34 & 1.4407 & \multirow{3}{*}{D20}  & Llama-3.1  & 27.05 & 73.40 & 1.0228 \\
                               & GPT-4o     & 24.47 & 70.36 & 1.1277 &                         & GPT-4o     & \textbf{37.23} & 82.52 & \textbf{0.8252} \\
                               & \cellcolor{gray!20}Ours  & \cellcolor{gray!20}\textbf{34.19} & \cellcolor{gray!20}\textbf{82.89} & \cellcolor{gray!20}\textbf{0.8526} &                         & \cellcolor{gray!20}Ours  & \cellcolor{gray!20}36.05 & \cellcolor{gray!20}\textbf{83.47} & \cellcolor{gray!20}0.8322 \\
        \bottomrule
    \end{tabular}
    \caption{Performance comparison across 20 evaluation criteria. Each dimension (Dim.) is evaluated using accuracy (ACC\textuparrow), soft accuracy ($\mathrm{ACC_{soft}}$\textuparrow), and the mean absolute error (MAE\textdownarrow).}
    \label{tab:performance_D1_D20}
\end{table*}

\section{Case Study}
\label{sec:case_study}

\begin{figure*}[h!]
  \centering
  \includegraphics[width=\linewidth]{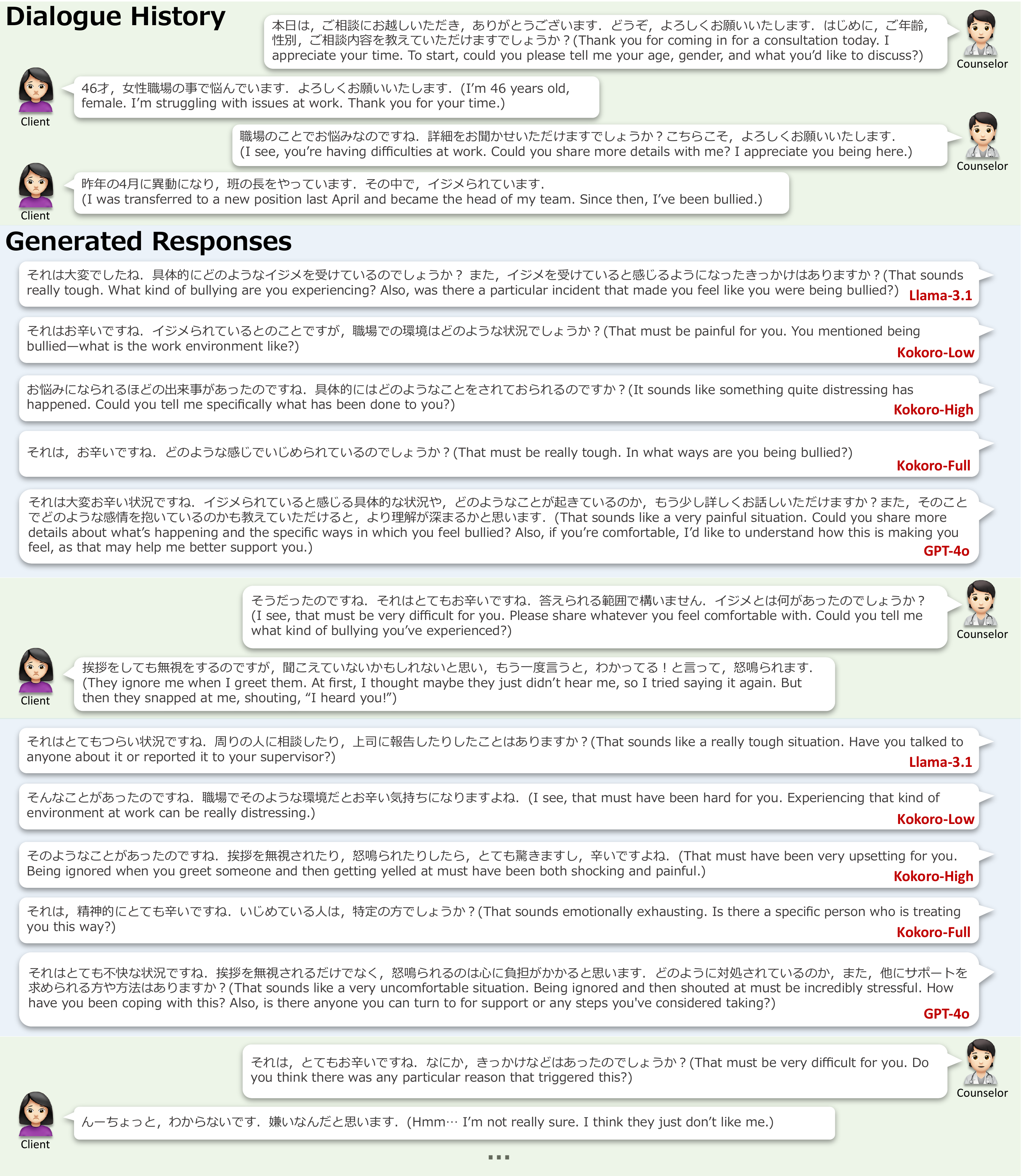}
  \caption{Examples of generated responses from each model.}
  \label{fig:fig_case_study}
\end{figure*}

Figure~\ref{fig:fig_case_study} presents example responses generated by each model. Although our model received a lower score than GPT-4o in human evaluations, its responses still demonstrated appropriate empathy, indicating a certain level of psychological counseling dialogue capability.

Additionally, we observed that models without fine-tuning (Llama-3.1 and GPT-4o) tend to frequently ask questions in their responses. However, in actual psychological counseling, not all situations require additional questioning to gather more information. On the contrary, when a client actively expresses themselves, providing only empathetic responses—without further questioning—can prevent unnecessary interruptions, facilitating smoother emotional expression and better conversation flow.

While this study’s human evaluation primarily focused on overall dialogue quality, it did not include a fine-grained analysis of question appropriateness. A well-structured questioning strategy is crucial in psychological counseling, as it can guide clients toward deeper reflection and help establish trust. However, excessive questioning may disrupt the conversation’s rhythm and even affect the client’s emotional stability.
Therefore, in future research, we aim to conduct more fine-grained human evaluations, assessing the appropriateness of questions posed by each model, among other factors.

\end{document}